\documentclass{isprs} 
\usepackage{subfigure}
\usepackage{setspace}
\usepackage{geometry} 
\usepackage{epstopdf}

\usepackage[labelsep=period]{caption}  
\usepackage[british]{babel} 
\usepackage{amsmath}
\usepackage[hang]{footmisc}

\usepackage{tabularx}
 \usepackage{booktabs}
 \usepackage{natbib} 
 \usepackage{siunitx}
 \usepackage{varwidth}
\usepackage{amsmath}
\usepackage{xcolor}
\usepackage{float}
\usepackage{tabularx,booktabs}
\usepackage{siunitx}
\usepackage{subfigure} 
\usepackage{url}
\usepackage{subfigure}

\begin{document}

\title{HoloGS: Instant Depth-based 3D Gaussian Splatting with Microsoft HoloLens 2}
\date{}


\author{
 Miriam Jäger\thanks{Corresponding author} , Theodor Kapler, Michael Feßenbecker, Felix Birkelbach, Markus Hillemann, Boris Jutzi}

\address{
	Institute of Photogrammetry and Remote Sensing (IPF), Karlsruhe Institute of Technology (KIT), Karlsruhe, Germany\\
 (miriam.jaeger, markus.hillemann, boris.jutzi)@kit.edu
}
\address{
	Institute of Photogrammetry and Remote Sensing (IPF), Karlsruhe Institute of Technology (KIT), Karlsruhe, Germany\\
 (miriam.jaeger, markus.hillemann, boris.jutzi)@kit.edu
}



\abstract{In the fields of photogrammetry, computer vision and computer graphics, the task of neural 3D scene reconstruction has led to the exploration of various techniques. Among these, 3D Gaussian Splatting stands out for its explicit representation of scenes using 3D Gaussians, making it appealing for tasks like 3D point cloud extraction and surface reconstruction. Motivated by its potential, we address the domain of 3D scene reconstruction, aiming to leverage the capabilities of the Microsoft HoloLens 2 for instant 3D Gaussian Splatting. We present HoloGS, a novel workflow utilizing HoloLens sensor data, which bypasses the need for pre-processing steps like Structure from Motion by instantly accessing the required input data i.e. the images, camera poses and the point cloud from depth sensing. We provide comprehensive investigations, including the training process and the rendering quality, assessed through the Peak Signal-to-Noise Ratio, and the geometric 3D accuracy of the densified point cloud from Gaussian centers, measured by Chamfer Distance. We evaluate our approach on two self-captured scenes: An outdoor scene of a cultural heritage statue and an indoor scene of a fine-structured plant. Our results show that the HoloLens data, including RGB images, corresponding camera poses, and depth sensing based point clouds to initialize the Gaussians, are suitable as input for 3D Gaussian Splatting.
}


\maketitle


\section{Introduction}\label{INTRODUCTION}

\sloppy

3D scene reconstruction is a fundamental task in the fields of computer vision, computer graphics and photogrammetry. Recently, however, methods have been gaining popularity that have the potential to revolutionize the classical workflows. This has been particularly initiated by the pioneering research on Neural Radiance Fields (NeRFs) \citep{mildenhall_et_al_2020}.
\paragraph{NeRFs and HoloLens.} NeRFs enable the rendering of novel views with the so-called view synthesis of a 3D scene in space with a neural network. The neural network is trained based on a set of images and corresponding camera poses and estimates a position-dependent density value and view-dependent RGB color values per position.
Through the volume density of points in a continuous space, geometries can be extracted. However, this requires techniques such as density thresholding or other methods of generating explicit 3D (surface) reconstructions from continuous neural network outputs \citep{unisurf, NeuS, volsdf, NeuralWarp, deepsdf, regSDF, neuralangelo, densitygradient_jaeger}, while the density carries an inherent uncertainty \citep{Jaeger_Nerfensemble}.
Most commonly, traditional methods like Structure from Motion (SfM) are used to calculate the interior orientation and the camera poses needed for training the NeRFs in a pre-processing step. 
As an alternative to SfM, the Microsoft HoloLens has proven to be an interesting interface, since it enables the extraction of the required input data, the images and corresponding poses \citep{hololens_nerf_jaeger}. Moreover, it has consistently demonstrated its efficacy as a mapping system (\citeauthor{jaeger}, \citeyear{jaeger}; \citeauthor{hololens_weinmann}, \citeyear{hololens_weinmann}, \citeauthor{hololens_iphone}, \citeyear{hololens_iphone}) and enables the real-time \citep{hololens_dennis}, highly detailed, colorized, 3D scene reconstruction and mobile mapping \citep{hololens_nerf_jaeger} with NeRFs.

\newpage
\textbf{Gaussian Splatting and HoloLens.} With regard to 3D scene reconstruction, particularly 3D Gaussian Splatting (GS) \citep{kerbl3Dgaussians} is outstanding due to its explicit representation of the scene utilizing 3D Gaussians. During optimization, these Gaussians are densified and adapted, undergoing growth, shrinkage, and adjustments in color and shape, until the photometric error between rendered images and training images becomes minimal. In contrast to the continuous radiance field representation of NeRF, these Gaussians explicitly represent the scene geometry, enabling direct access to it. When it comes to photogrammetry and 3D computer vision, this is particularly of interest for 3D point cloud extraction and surface reconstruction.
In contrast to most NeRF methods, further input data in the form of a point cloud is required for 3D Gaussian Splatting, for which the sparse point cloud from SfM is usually used. This point cloud is used to initialize the Gaussians. Thus, pre-processing and calculation steps are required to compute the camera poses and sparse point cloud from the images. At this point, the Microsoft HoloLens once again becomes relevant. Alongside the RGB images and their corresponding camera poses, the HoloLens provides depth images and corresponding camera poses, which can be transformed into the required point cloud. This enables instant 3D scene reconstruction with 3D Gaussian Splatting, i.e. without additional time-consuming pre-processing steps.

\begin{figure}[h]
\begin{center}
  \includegraphics[width=1.0\columnwidth]{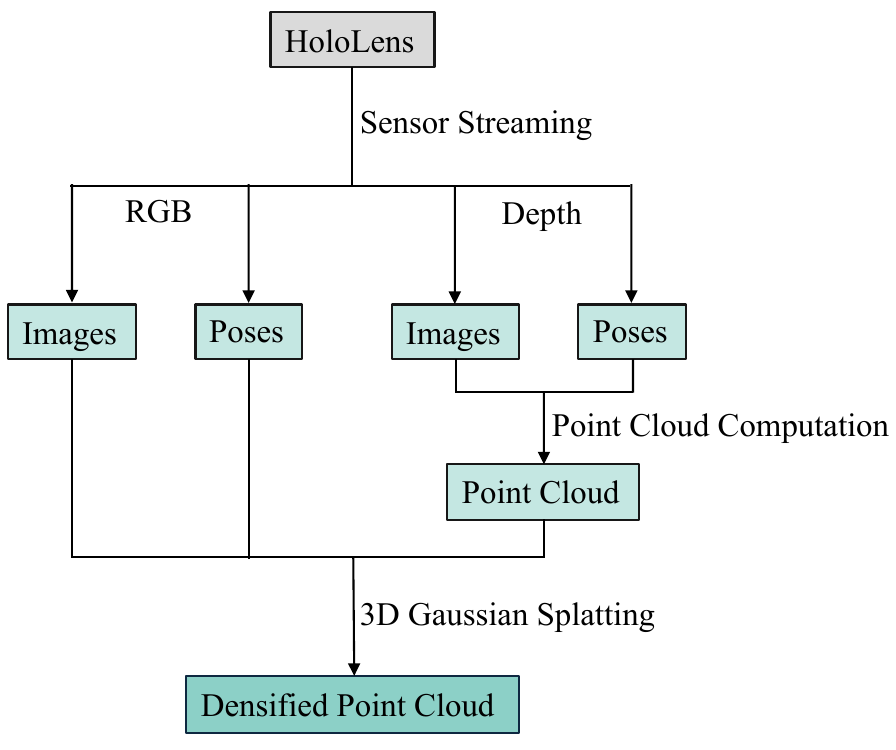}
	\vspace{-3mm}
	\caption{Flowchart of HoloGS: Via the HoloLens 2 sensor streaming, the required data is directly extracted and processed during data capturing. From the depth data, a point cloud is instant created which, together with the RGB images and corresponding camera poses, is then fed into 3D Gaussian Splatting.}
\label{fig:flowchart}
\end{center}
\end{figure}

In this work, we present HoloGS (Figure \ref{fig:flowchart}), for an instant 3D scene reconstruction by 3D Gaussian Splatting with Microsoft HoloLens 2 data. This is done directly based on sensor information, since the HoloLens enables to access the required input data, i.e. the images, camera poses and the point cloud from depth sensing in real-time. We investigate whether the data quality of the HoloLens is sufficient for 3D Gaussian Splatting. In order to evaluate our workflow, we additionally follow the traditional pipeline, which uses COLMAP \citep{Schonberger_2016_CVPR}, to estimate the camera poses and the sparse point cloud. Our analysis includes examining the training process and evaluating the rendering quality of our results by using the Peak Signal-to-Noise Ratio (PSNR) and photometric loss. Furthermore, we report the geometric 3D accuracy quantitatively as well as qualitatively of the resulting densified point clouds from the Gaussian centers using cloud-to-cloud Chamfer Distance. Thereby, we envision a refilling of the (sparse) input point cloud, comparable to a post-processing step by Multi-View Stereo.

We demonstrate that HoloGS with Microsoft HoloLens 2 data, comprising the RGB images with corresponding camera poses, and the point cloud from depth sensing, is suitable as input for 3D Gaussian Splatting. The rendered images reasonably reflect the geometry and appearance. Furthermore, HoloGS enables refilling by the extraction of a densified point cloud from Gaussian centers.

\section{Methodology}\label{sec:METHOD}

Section \ref{sec:poses_pc_sfm_hololens} outlines the principles of the methods used to determine the input data for 3D Gaussian Splatting: via the standard method with external data from SfM and via our approach for instant 3D Gaussian Splatting with internal data from Microsoft HoloLens 2. Subsequently, Section \ref{sec:implementation} presents the implementation details for Gaussian Splatting.
Finally, Section \ref{sec:method_recon} describes our method for extracting the densified point cloud from Gaussian Splatting after the training process.

\subsection{Initialization} \label{sec:poses_pc_sfm_hololens}

\paragraph{External SfM data.}

As mentioned, the standard workflow uses SfM to determine the camera poses and the sparse point cloud in a pre-processing step. SfM in general describes the procedure of the reconstruction of a 3D scene from a set of images, which taken from different directions and positions by a camera in motion. It relies on the calculation and matching of point correspondences within an image sequence from overlapping images. Most commonly by using methods such as SIFT \citep{Lowe}. The resulting products are the camera poses, the camera intrinsics as well as a sparse point cloud from the point correspondences.
In addition, the fundamental SfM sparse point cloud allows a following Multi-View Stereo (MVS) pipeline \citep{Schonberger_2016_CVPR} regarding a dense reconstruction, which densifies the SfM point cloud. We consider the MVS dense reconstruction as a reference point cloud for comparison.
In this paper, the (incremental) SfM technique by \citep{Schonberger_2016_CVPR} from the original implementation of 3D Gaussian Splatting \footnote{\url{https://github.com/graphdeco-inria/gaussian-splatting} (last access 12/01/2023)} is used for the external data calculation of the camera poses and the sparse point cloud. This process is conducted using the same HoloLens RGB images to ensure uniform conditions similar to those of the following internal data approach.

\paragraph{Internal HoloLens data.}
For an instant 3D scene reconstruction with 3D Gaussian Splatting directly from Microsoft HoloLens 2, HoloGS targets three main steps, according to Figure \ref{fig:flowchart}: Sensor streaming, real-time point cloud computation, and instant 3D Gaussian Splatting.
The HoloLens 2 server application \citep{HoloLens2_Streaming} is used for requesting the data in the Microsoft HoloLens 2\footnote{\url{https://www.microsoft.com/en-us/hololens/hardware} \\(last access 02/08/2024)}. The system provides access to all of the HoloLens 2 sensors, including the RGB images from the 1920\,$\times$\,1080 camera, interior orientation (camera intrinsics) and corresponding camera poses, as well as the depth sensor for depth images and corresponding poses. 
Firstly, through the sensor streaming of the HoloLens, the RGB images and corresponding camera poses including the interior orientation as well as the depth images and their camera poses are queried and extracted from the sensor system during data acquisition.
Secondly, the depth images are each transformed into a 3D point cloud by calculating the corresponding 3D point for each pixel in the depth image based on the interior orientation of the depth camera and the depth information. These point clouds are subsequently merged into a joint point cloud via the camera poses.
Lastly, the required data, i.e., the RGB images with corresponding camera poses and the point cloud from the depth information transformed to the coordinate system of the RGB camera, is fed to 3D Gaussian Splatting as initial data. In this context, the RGB images serve as training data for optimizing the Gaussians by minimizing the photometric error between rendered images and their real counterparts at the same camera poses. The 3D points of the point cloud from the depth information forms the centers of the initial Gaussians. The other parameters of the Gaussians are initialized as in the original implementation of \citeauthor{kerbl3Dgaussians}, i.e. isotropic Gaussians with axes equal to the mean of the distance to the closest three points.

\subsection{3D Gaussian Splatting Implementation}\label{sec:implementation}
After initialization, 3D Gaussian Splatting is processed according to the original implementation. We train on the default parameters with learning rates of 0.0025 for spherical harmonics features, 0.05 for opacity adjustments, 0.005 for scaling operations and 0.001 for rotation transformations, while the training incorporates 30\,000 iterations on a NVIDIA RTX3090 GPU. The photometric loss for the optimization is given by the following loss function \ref{equ:loss}.
\begin{equation}\label{equ:loss}
L = (1 - \lambda) L_{\text{1}} + \lambda L_{\text{D-SSIM}} \\
\end{equation}
with $\lambda$ = 0.2 by default, $L_{\text{1}}$-Norm of the per pixel color difference and $L_{\text{D-SSIM}}$-Term \citep{kerbl3Dgaussians}.

\subsection{3D Point Cloud Extraction}\label{sec:method_recon}

As mentioned above, the point clouds from the external SfM or from the internal depth information of the HoloLens serve as initial Gaussians.
Through the optimization process of Gaussian Splatting, additional points are generated based on color information of the images. In doing so, Gaussians grow, split, shrink or are removed.
Based on this, we envision a refilling of the sparse input point cloud, comparable to a post-processing step by MVS. Especially, since the SfM sparse point cloud in particular is only created based on the features from SIFT point correspondences, it is, as the name implies, relatively sparse.
After the training of 3D Gaussian Splatting, the densified and optimized point cloud can be extracted. On the assumption that 3D information containing color also exists as actual geometry in the scene, we consider the centers of the Gaussians, which represent each the mean of each Gaussian, as 3D geometry in our approach.

\section{Dataset}\label{sec:dataset}
\begin{figure}[h!]
	\centering
  \hspace{-2mm}
\subfigure[]{\label{fig:CapturingData}
	\includegraphics[width=0.40\linewidth]{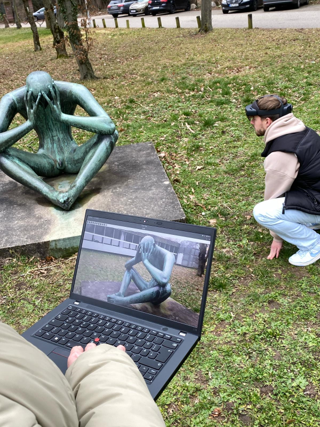}}
 \hspace{-2mm}
	\subfigure[]{\label{fig:denker_poses}
	\includegraphics[width=0.50\linewidth]{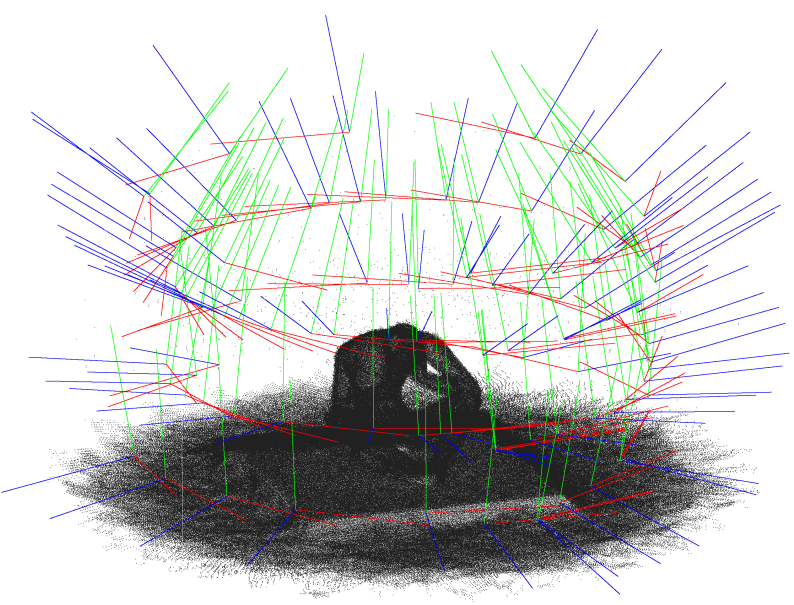}}
	\vspace{-3mm}
	\caption{\protect{\subref{fig:CapturingData}} Data capturing with Microsoft HoloLens 2 and its streaming application \citep{HoloLens2_Streaming}. \protect{\subref{fig:denker_poses}} Point cloud based on depth data of the scene 'Denker' and camera poses visualized by colored coordinate frames.}
\label{fig:Datenaufnahme}
\end{figure}

\begin{figure}[h!]
	\centering
 	\subfigure[]{\label{fig:homo_input_sfm}
	\includegraphics[width=0.45\linewidth]{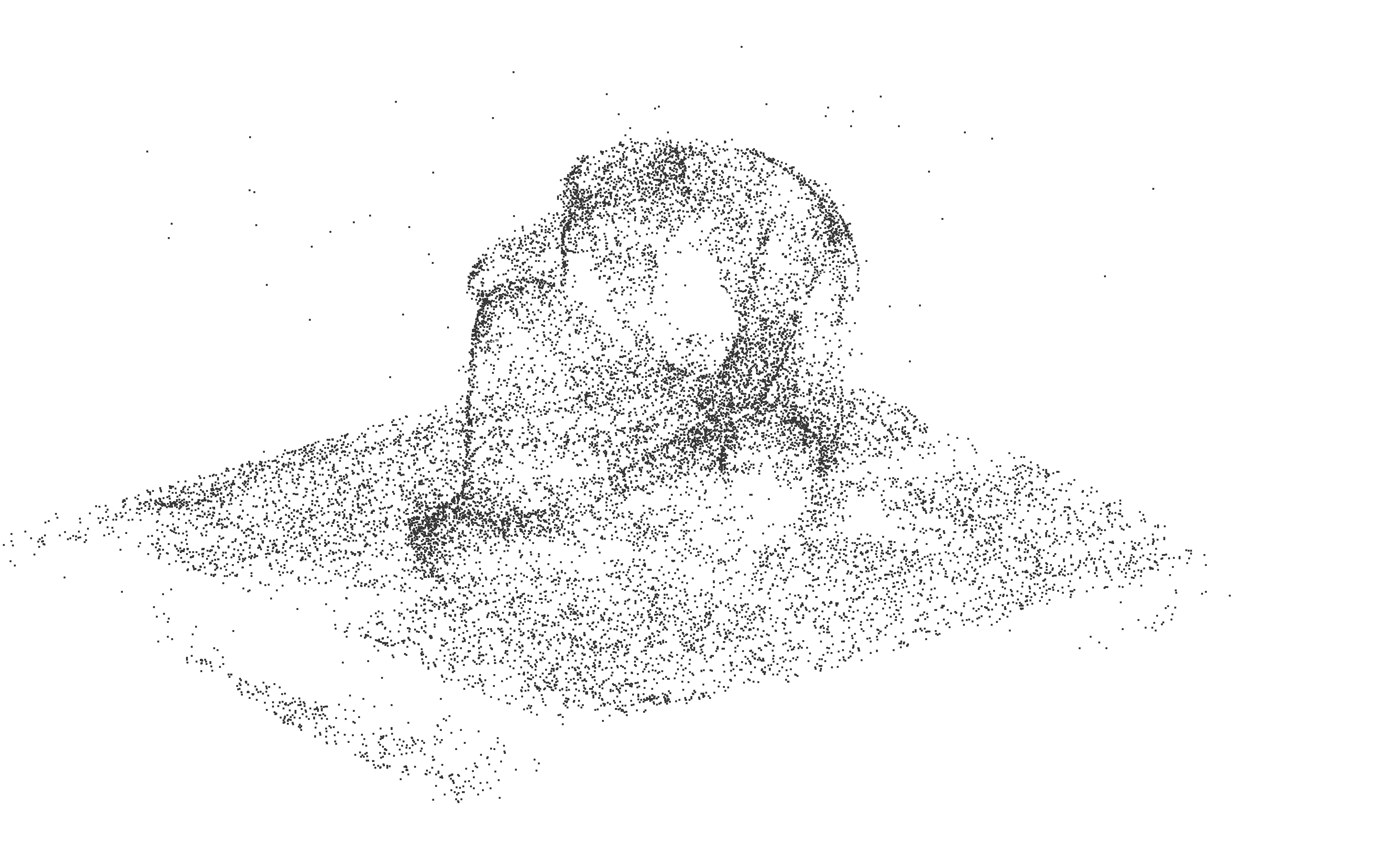}}
	\hspace{-2mm}
\subfigure[]{\label{fig:homo_input_hololens}
	\includegraphics[width=0.45\linewidth]{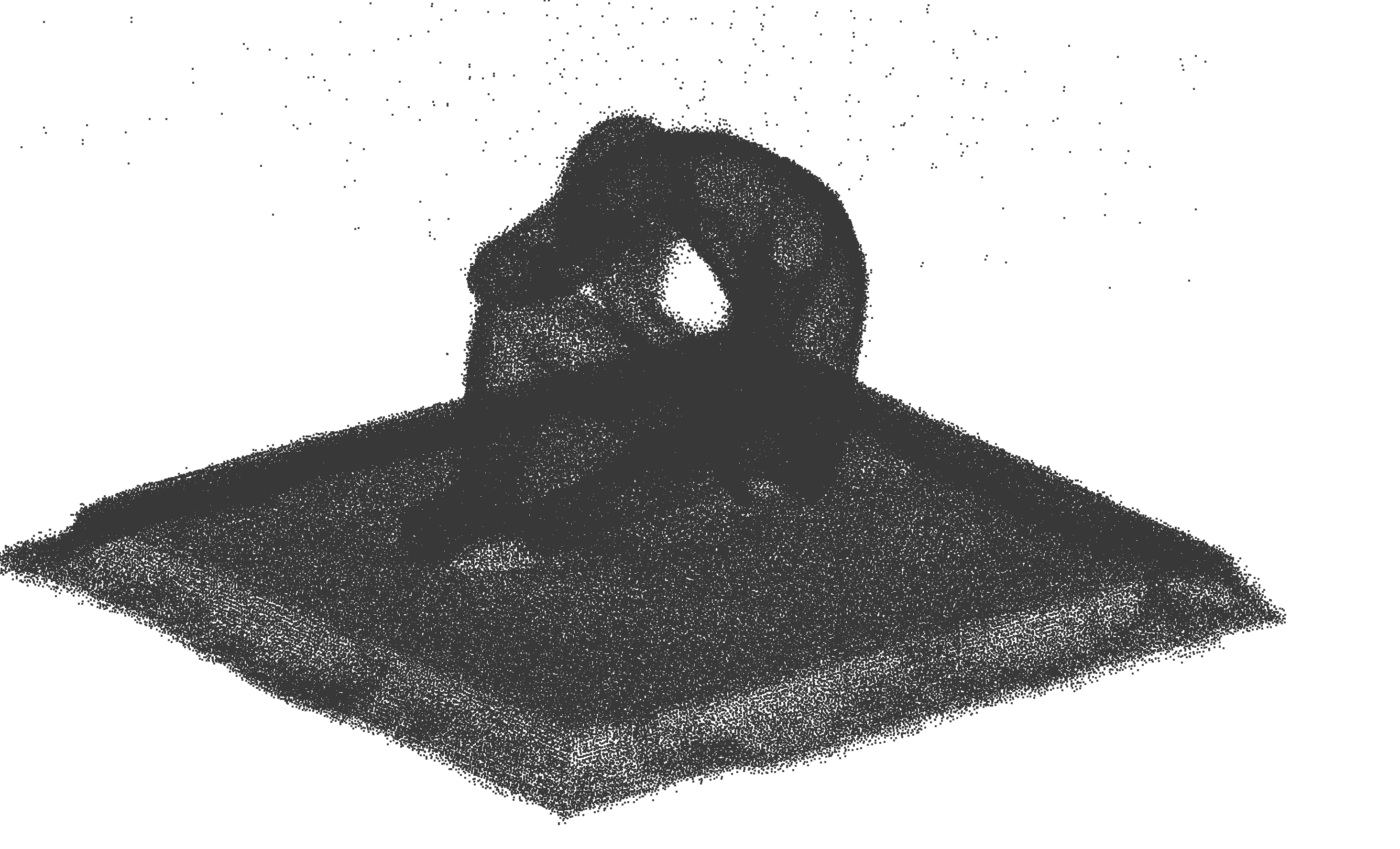}}
	\vspace{-3mm}
	\caption{Initialization input 3D point clouds. \protect{\subref{fig:homo_input_sfm}} sparse point cloud from SfM and
 \protect{\subref{fig:homo_input_hololens}} point cloud calculated based on the depth images from HoloLens.}
\label{fig:input_point_clouds}
\end{figure}

Our experiments are based on two datasets that we captured with Microsoft HoloLens 2: An outdoor scene of a cultural heritage statue, called 'Denker' (Figure \ref{fig:Datenaufnahme}), and an indoor scene, called 'Ficus' of a fine-structured plant. Both datasets contain 111 images each of size 1920\,$\times$\,1080 with camera poses, captured in a hemispherical camera framing, as well as the respective depth maps and poses, whereby we mask out depth information above 3m.
The respective initialization type (Section \ref{sec:poses_pc_sfm_hololens}) of the input data results in different point clouds: The SfM sparse point cloud (Figure \ref{fig:homo_input_sfm}) including over 40\,000 points, and the point cloud (Figure \ref{fig:homo_input_hololens}) from the depth images from HoloLens including over 620\,000 points .

\section{Experiments and Results}\label{RESULTS}
In this section, we present our experiments and results by a quantitative evaluation on analyzing the training process by rendering quality in Section \ref{sec:training}. This is followed by a qualitative analysis of the rendered images in Section \ref{sec:rendering}. Finally, we evaluate the geometric 3D reconstruction based on the densified point clouds from the Gaussian centers quantitative as well as qualitative in Section \ref{sec:reconstruction}.

\subsection{Training}\label{sec:training}
\begin{figure}[H]
	\centering
\subfigure[]{\label{fig:denker}
\hspace{-4mm}
	\includegraphics[width=1.05\columnwidth]{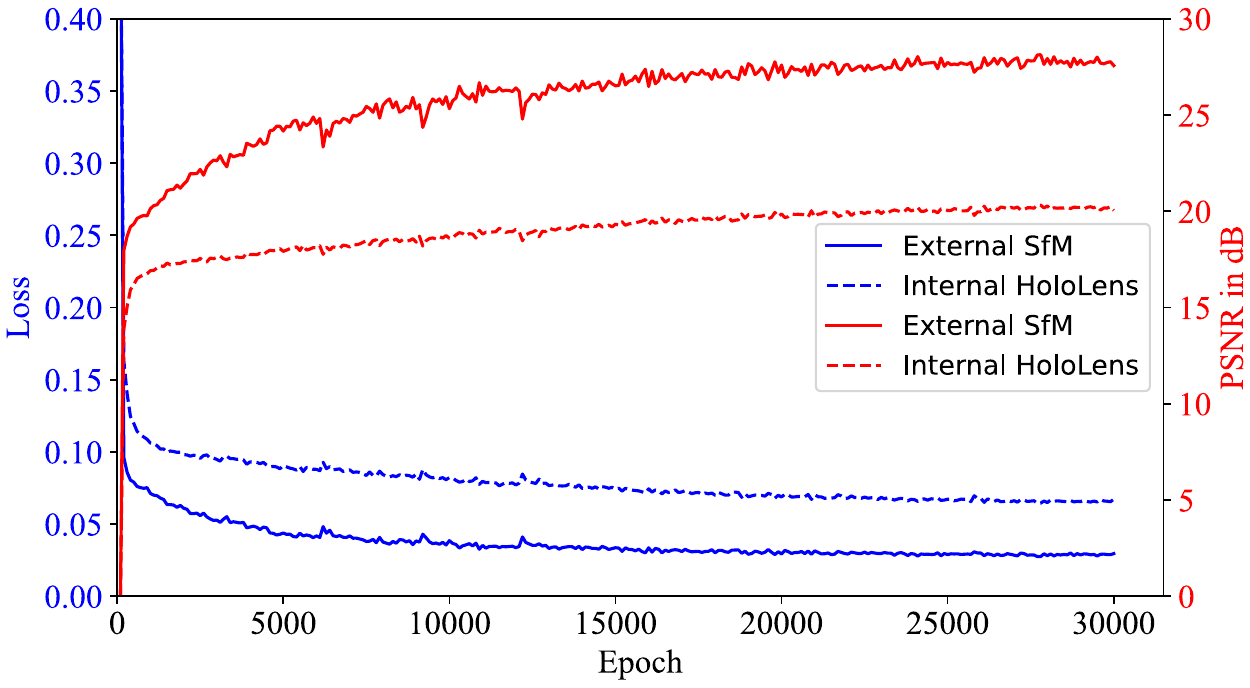}} \\
	\vspace{-2.5mm}
\subfigure[]{\label{fig:ficus} 
	\hspace{-4mm}
     \includegraphics[width=1.05\columnwidth]{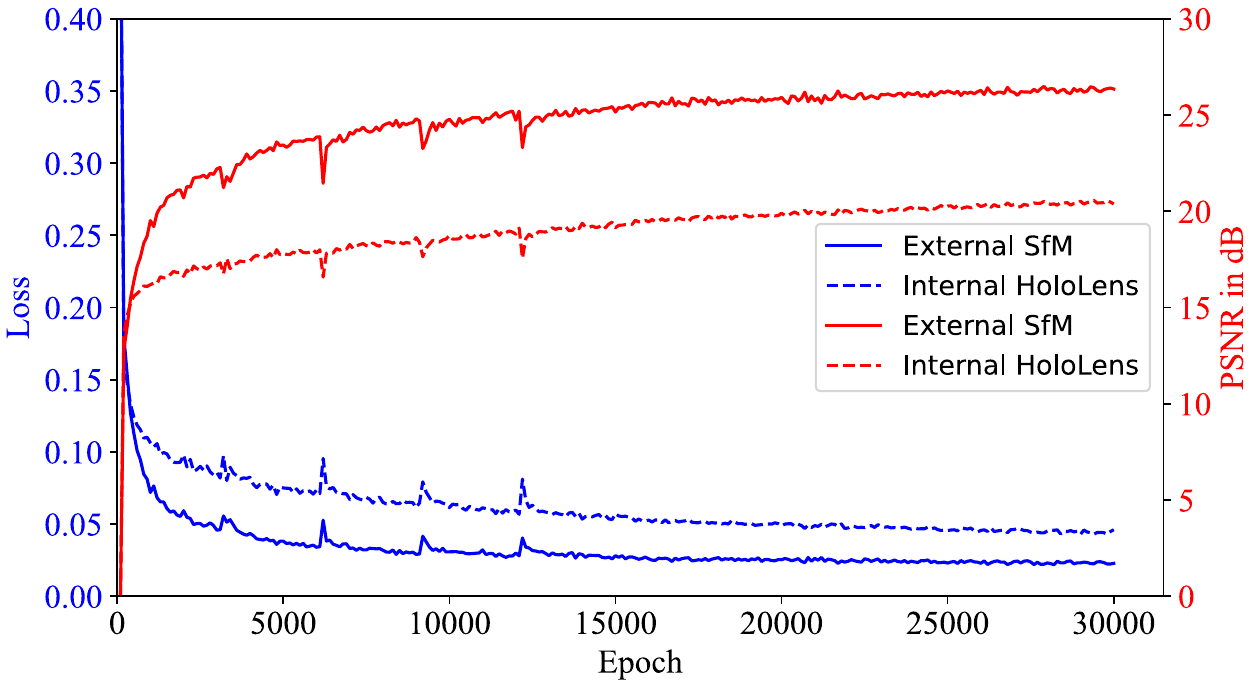}}
     	\vspace{-3mm}
	\caption{Comparison of the Peak Signal-to-Noise Ratio (PSNR) $\uparrow$ in \si{\dB} and loss $\downarrow$ during the training processes with 30\,000 iterations with 3D Gaussian Splatting with different types of input data. Top: \protect{\subref{fig:denker}} external SfM and internal HoloLens data on scene 'Denker'. Bottom: \protect{\subref{fig:ficus}} external SfM and internal HoloLens data on scene 'Ficus'. The red curves show the PSNR, the blue curves the training loss. }
\label{fig:Training}
\end{figure}

We evaluate the training process with the Peak Signal-to-Noise Ratio (PSNR) \citep{mildenhall_et_al_2020}, which is a common metric in NeRF context. Figure \ref{fig:Training} shows the change in PSNR and training loss \citep{kerbl3Dgaussians} (Equation \ref{equ:loss}) over the iterations. It demonstrates that HoloGS with internal HoloLens data, including RGB images, corresponding camera poses, and point clouds derived from the depth map, leads to relatively smooth convergence of 3D Gaussian Splatting. Convergence occurs after approximately 25,000 iterations, reaching a maximum PSNR (Table \ref{tab:psnr}) of around 20.55 \si{\dB} for the scene 'Denker' and 20.17 \si{\dB} for the scene 'Ficus'. Notably, the convergence is rapidly achieved with the HoloLens data for both scenes.
In contrast, utilizing external SfM data yields higher PSNR values of 27.54 \si{\dB} for the scene 'Denker' and 26.21 \si{\dB} for the scene 'Ficus'.
Conversely, the loss for the internal HoloLens data during training is higher than for the external SfM data for both scenes. Interestingly, the curves show peaks every 3000 iterations up to iteration 15000. These peaks can be explained by the density moderation technique of Gaussian Splatting. This technique sets the opacity values close to zero every 3000 iterations to prevent the method from getting stuck with floaters close to the camera poses which could cause an unjustified increase in the density of Gaussians \citep{kerbl3Dgaussians}.

\begin{table}[H]
\centering
\begin{tabular}{|l|c|c|}
\hline
& External SfM data & Internal HoloLens data\\
\hline
Denker & 27.54 & 20.17 \\
Ficus & 26.21 & 20.55 \\
\hline
\end{tabular}
    \vspace{-0.2cm}
\caption{Peak Signal-to-Noise Ratio (PSNR) $\uparrow$ in dB after 30\,000 iterations for the scenes 'Denker' and 'Ficus' each with external SfM and internal HoloLens data.}
\label{tab:psnr}
\end{table}

\subsection{Rendering Quality}\label{sec:rendering}
\begin{figure*}[h!]
	\centering
\subfigure[]{\label{fig:denker_without_dark_colmap_}  
     \includegraphics[width=0.44\columnwidth]{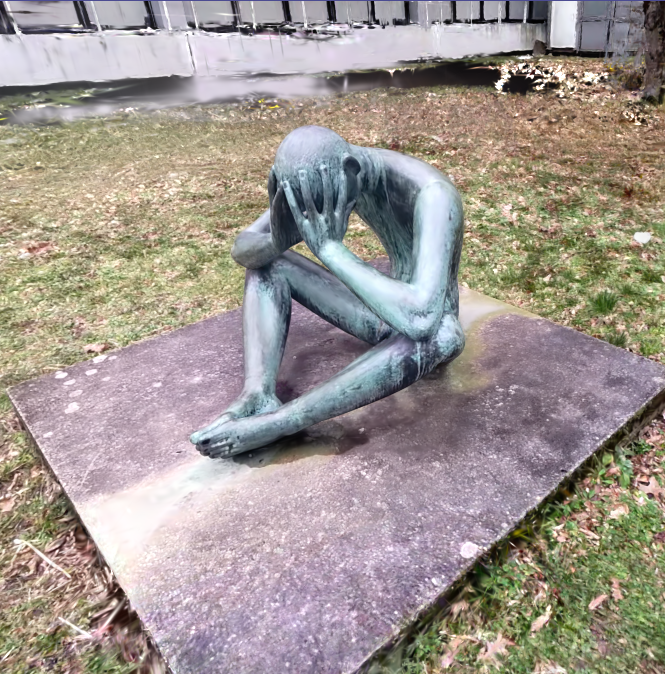}}
     	  \hspace{0.075cm}
     \subfigure[]{\label{fig:denker_without_dark}
	\includegraphics[width=0.445\columnwidth]{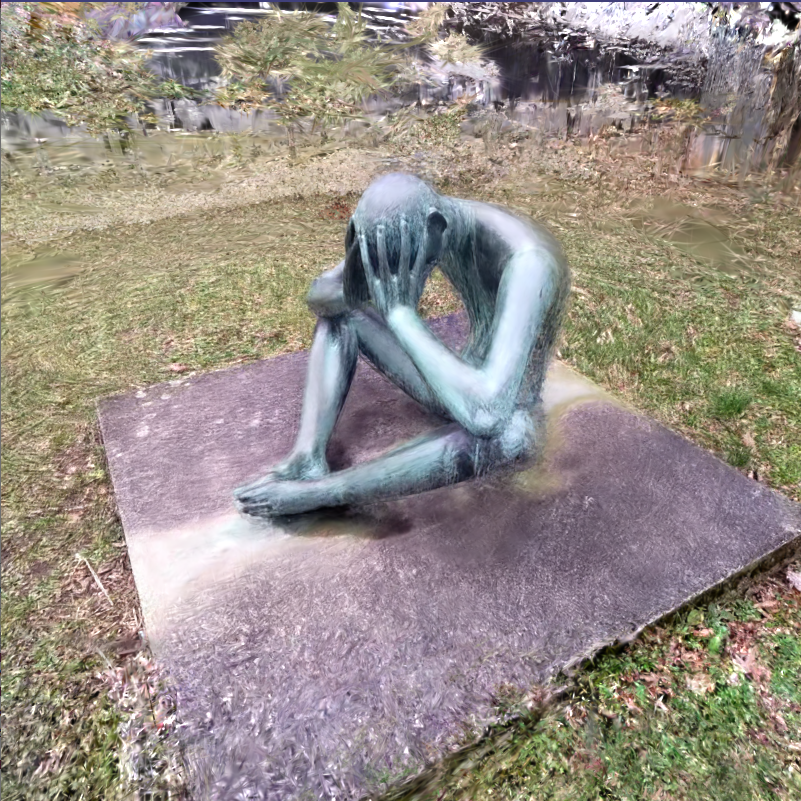}}
     \hspace{0.2cm}
\subfigure[]{\label{fig:pflanze_colmap_}  
     \includegraphics[width=0.495\columnwidth]{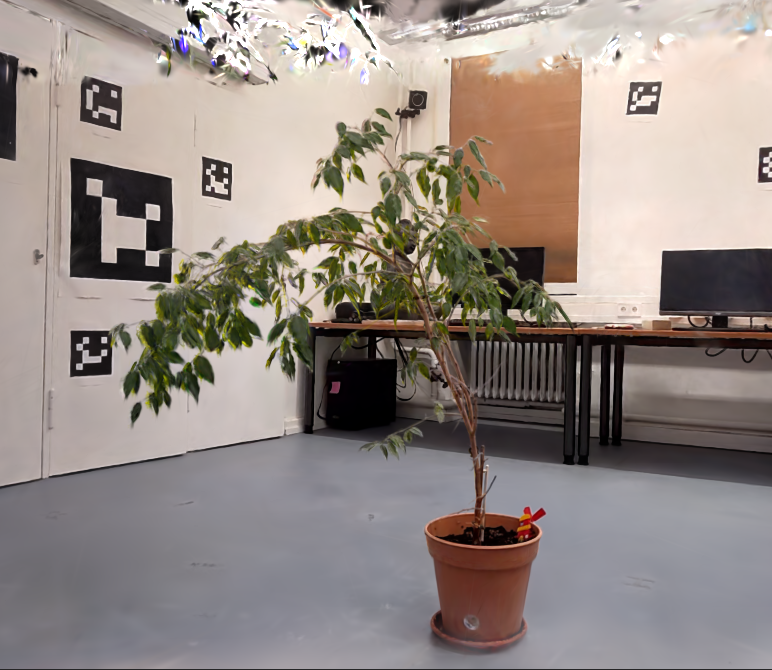}}
     	 \hspace{0.075cm}
          \subfigure[]{\label{fig:pflanze}
	\includegraphics[width=0.528\columnwidth]{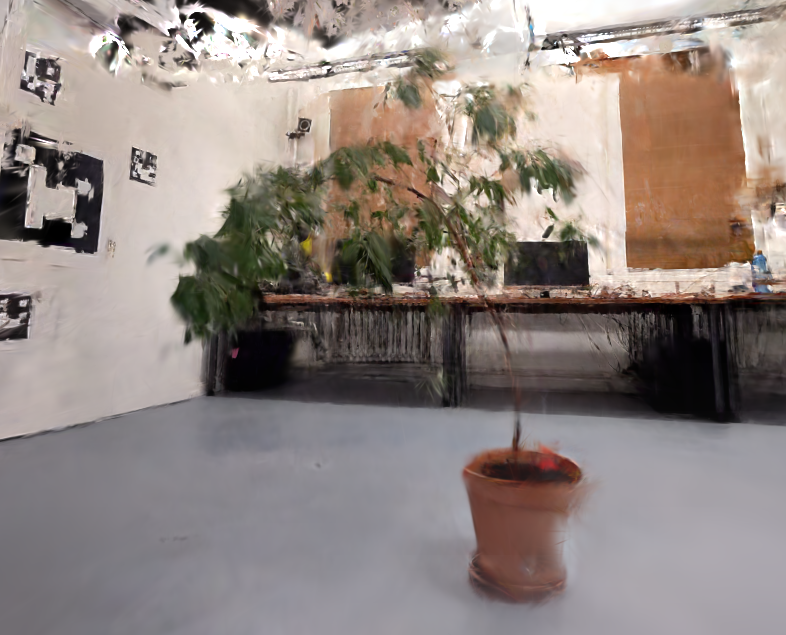}}
     \vspace{-0.35cm}
	\caption{Rendered images. From left to right: \protect{\subref{fig:denker_without_dark_colmap_}} external SfM data and \protect{\subref{fig:denker_without_dark}} internal HoloLens data on scene 'Denker', as well as \protect{\subref{fig:pflanze_colmap_}} external SfM data and \protect{\subref{fig:pflanze}} internal HoloLens data on scene 'Ficus'.}
\label{fig:Rendered}
\end{figure*}

The results of the rendered images in Figure \ref{fig:Rendered} closely correspond to the numerical results obtained during the training process. Particularly, the external SfM data computed during pre-processing demonstrates significantly improved performance. For the scene 'Denker', HoloGS produces satisfactory results for the statue itself, comparable to those derived from SfM data. Furthermore, in scene 'Ficus', HoloGS struggles to accurately represent the fine structures of the plant, leading to noise. Overall, HoloGS generally results in blurry edges of objects in the scene. In addition, large, foggy and blurry floater artifacts can be recognized in these scenes. In contrast, for the external SfM data, these artifacts are only evident in a limited number of areas, e.g. above the head of the statue in the scene 'Denker' and in the unobserved area of the scene 'Ficus' on the ceiling.

\subsection{3D Point Cloud Extraction}\label{sec:reconstruction}
\begin{figure*}[h!]
	\centering
\subfigure[]{\label{fig:Denker_SfM}
	\includegraphics[width=0.49\columnwidth]{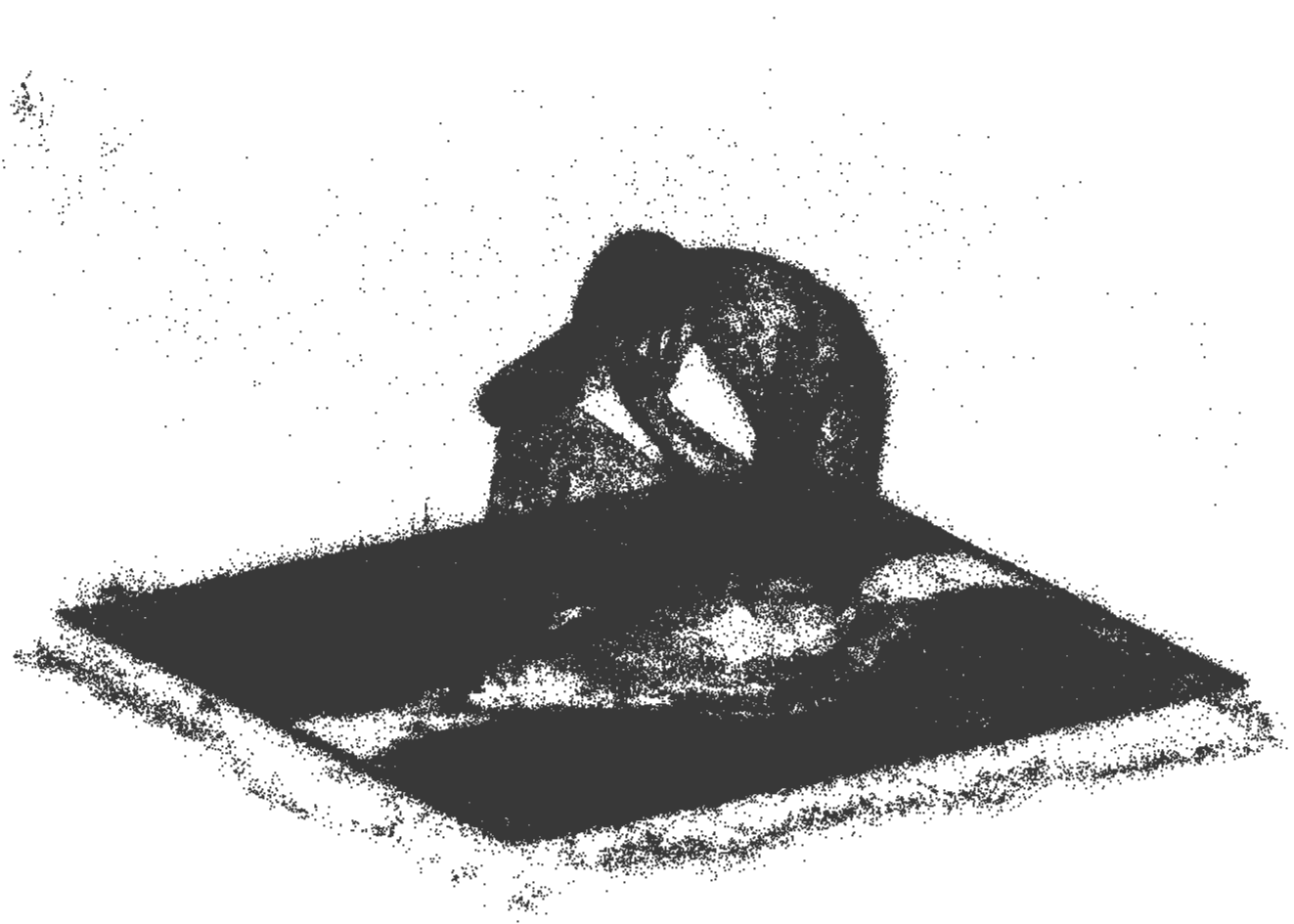}}
\subfigure[]{\label{fig:Denker_HoloLens}  
     \includegraphics[width=0.49\columnwidth]{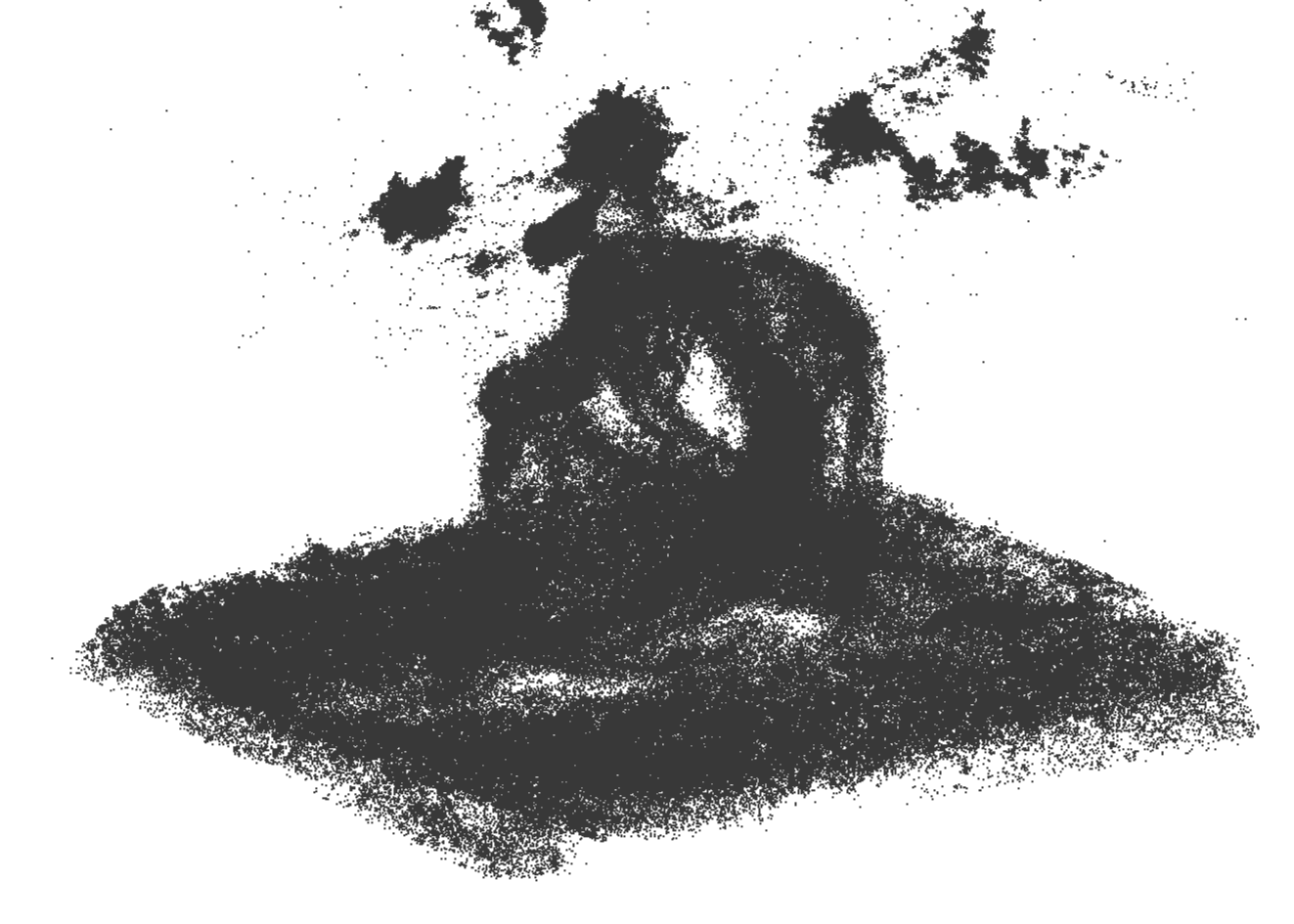}}
     \subfigure[]{\label{fig:Ficus_SfM_}
	\includegraphics[width=0.49\columnwidth]{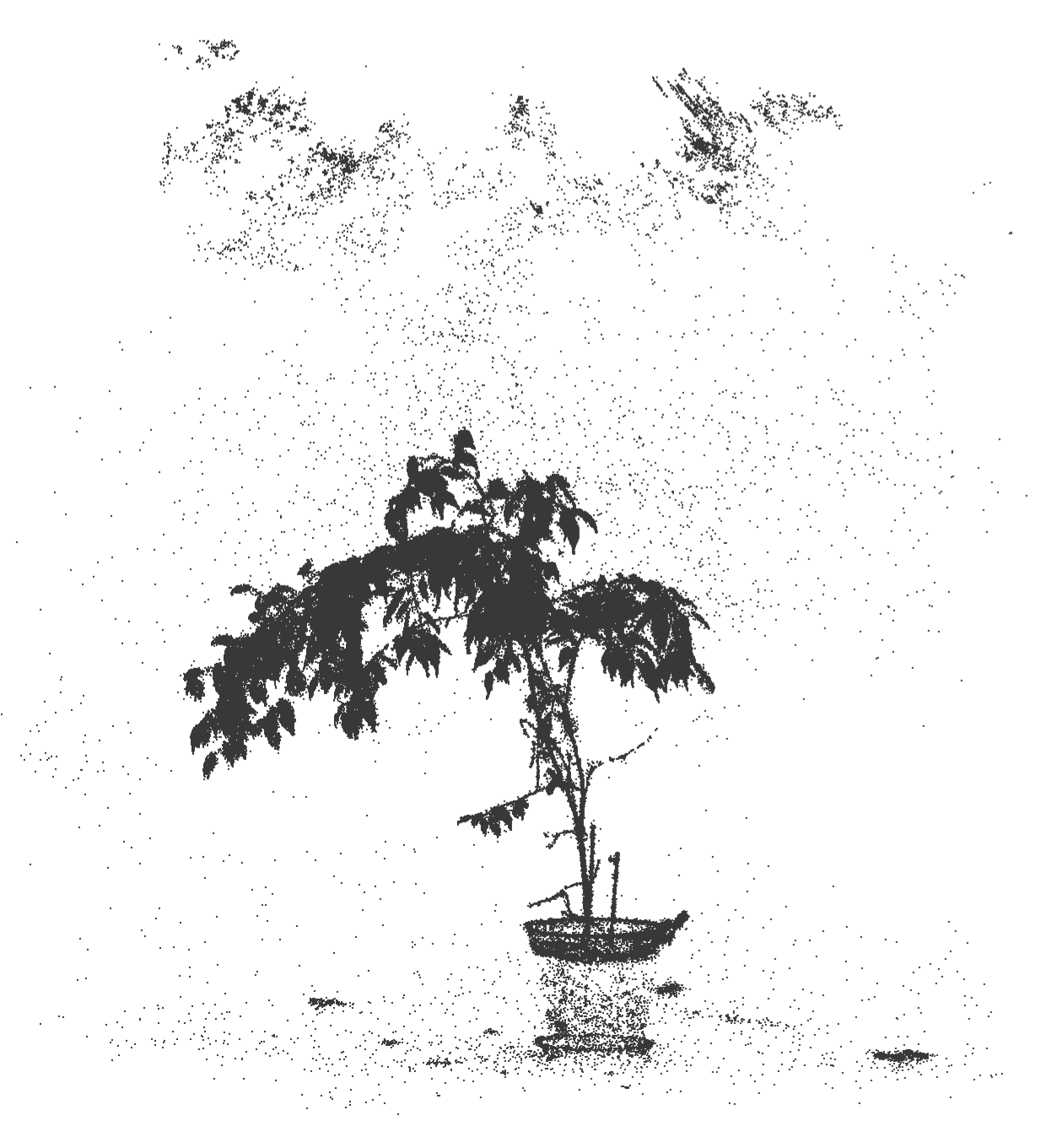}}
\subfigure[]{\label{fig:Ficus_HoloLens_}  
     \includegraphics[width=0.49\columnwidth]{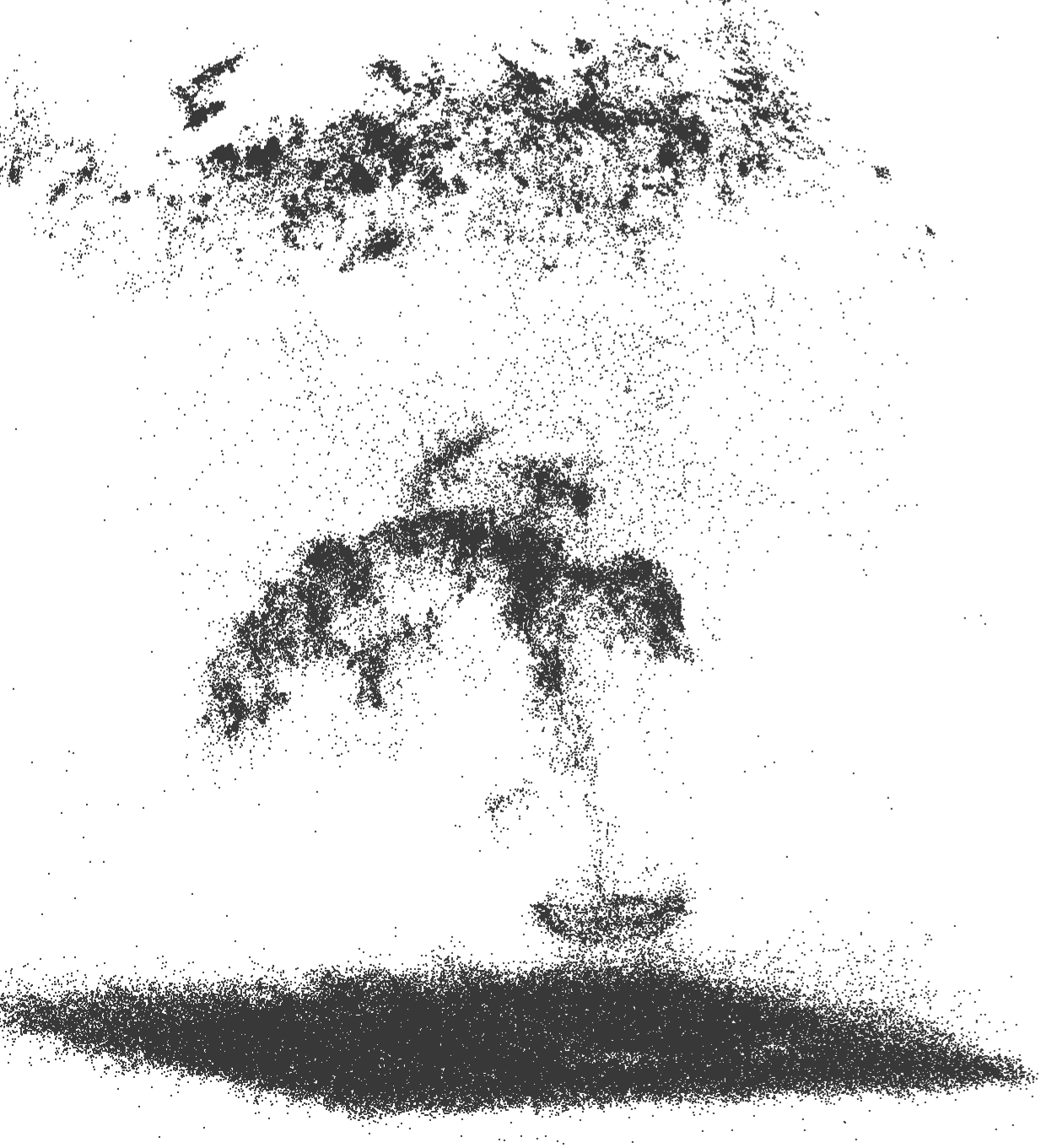}}
      \vspace{-0.35cm}
	\caption{Point clouds by extracting the center of each Gaussian after training process. From left to right: \protect{\subref{fig:Denker_SfM}} external SfM data on scene 'Denker', \protect{\subref{fig:Denker_HoloLens}} internal HoloLens data on scene 'Denker', \protect{\subref{fig:Ficus_SfM_}} external SfM data on scene 'Ficus', \protect{\subref{fig:Ficus_HoloLens_}} internal HoloLens data on scene 'Ficus'.}
\label{fig:Extracted_pointclouds}
\end{figure*}

We extract the densified point cloud, which is generated during the training process like described in Section \ref{sec:method_recon}. The extracted densified point clouds of the Gaussians (Figure \ref{fig:Extracted_pointclouds}) illustrate the differences in Gaussian Splatting for 3D scene reconstruction between the external SfM data and the internal HoloLens data.
For the scene 'Denker', there is an overall good coverage with the SfM data, where the structure of the object is clearly visible with sharp edges. Nonetheless, some gaps in the point cloud are evident, particularly on the platform and on the statue's arms and legs. In these areas the color differs from the rest and appears more homogeneous and low-textured. When using the internal HoloLens data, there is an identifiable structure of the object through the centers of the Gaussians. Although, the point cloud is overall noisy, with indistinct and fuzzy object edges. Additionally, large artifacts are present in the point clouds, corresponding to the floater artifacts in the rendered images.
For the scene 'Ficus', a similar pattern emerges. The point cloud from external SfM data exhibits clear and sharp edges, accurately capturing the fine structure of the vegetation of the plant. However, gaps are noticeable, especially in the pot area, characterized by a uniform, low-textured color. Additionally, small floater artifacts appear intermittently above the object, particularly in areas with lower scene coverage from the captures. When using the HoloLens data, the structure of the object is also clearly recognizable, with a high coverage. Nonetheless, the point cloud remains heavily noisy, with numerous floater artifacts, especially in areas above the object. Again, the low-textured pot of the plant is not full reconstructed.

\begin{figure*}[h!]
	\centering
\hspace{-1.75cm}
\rotatebox{90}{external SfM data}
\hspace{0.75cm}
\subfigure[]{\label{fig:Denker_Reference_a}
	\includegraphics[width=0.26\linewidth]{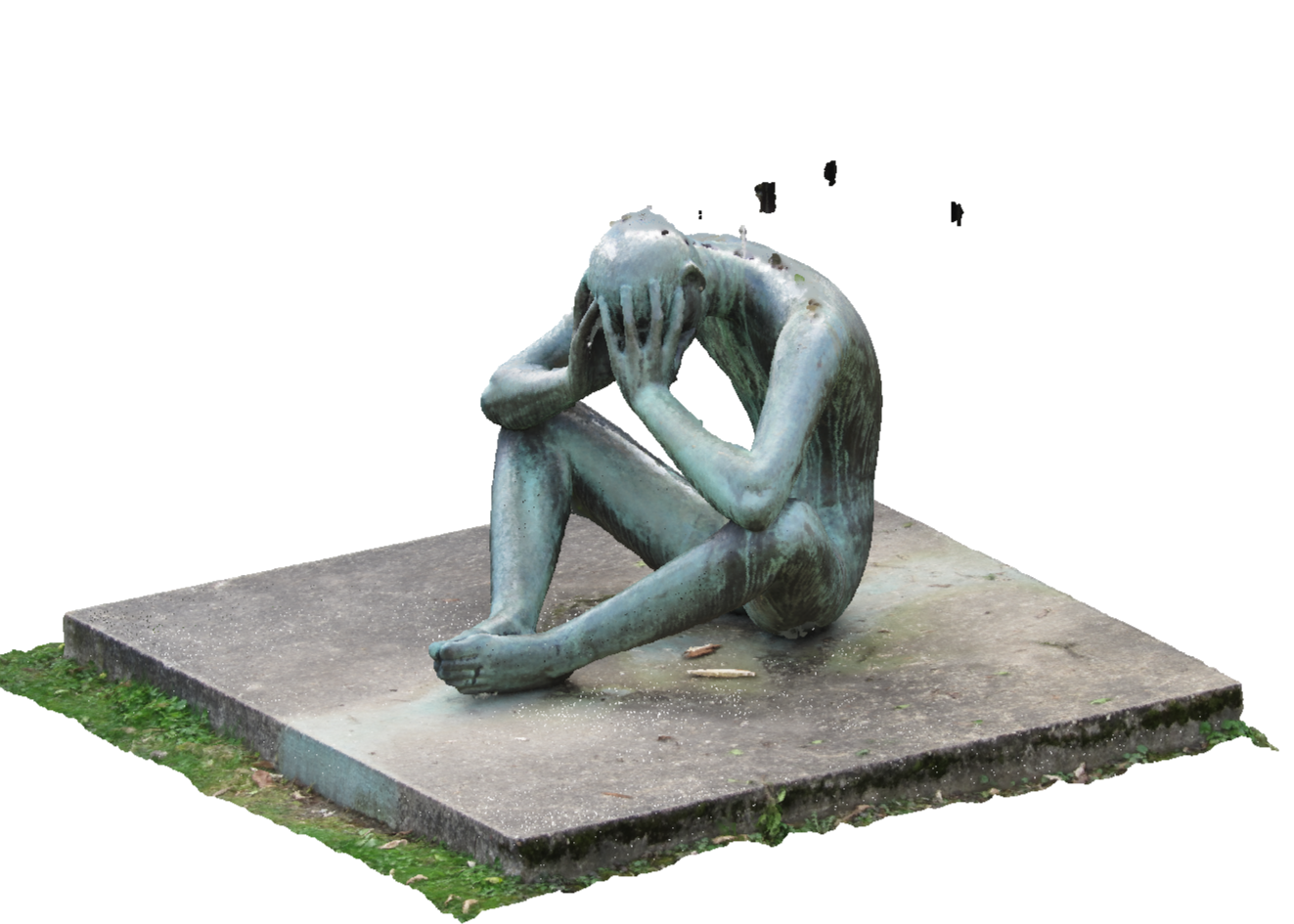}}
	\hspace{0.2cm}
\subfigure[]{\label{fig:Denker_Reference_SfM}
	\includegraphics[width=0.26\linewidth]{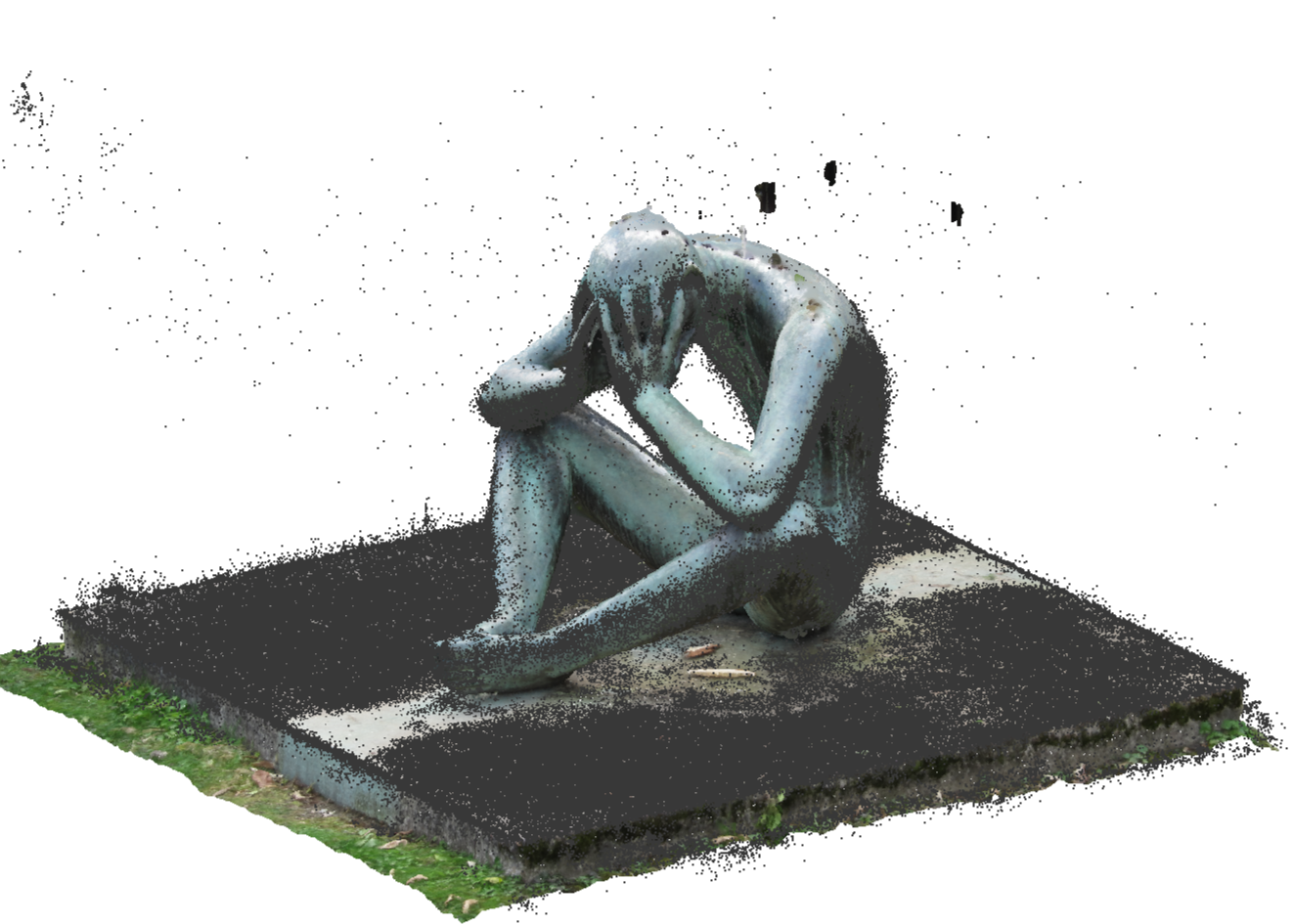}}
	\hspace{0.2cm}
\subfigure[]{\label{fig:Denker_SfM_C2C}  
     \includegraphics[width=0.26\linewidth]{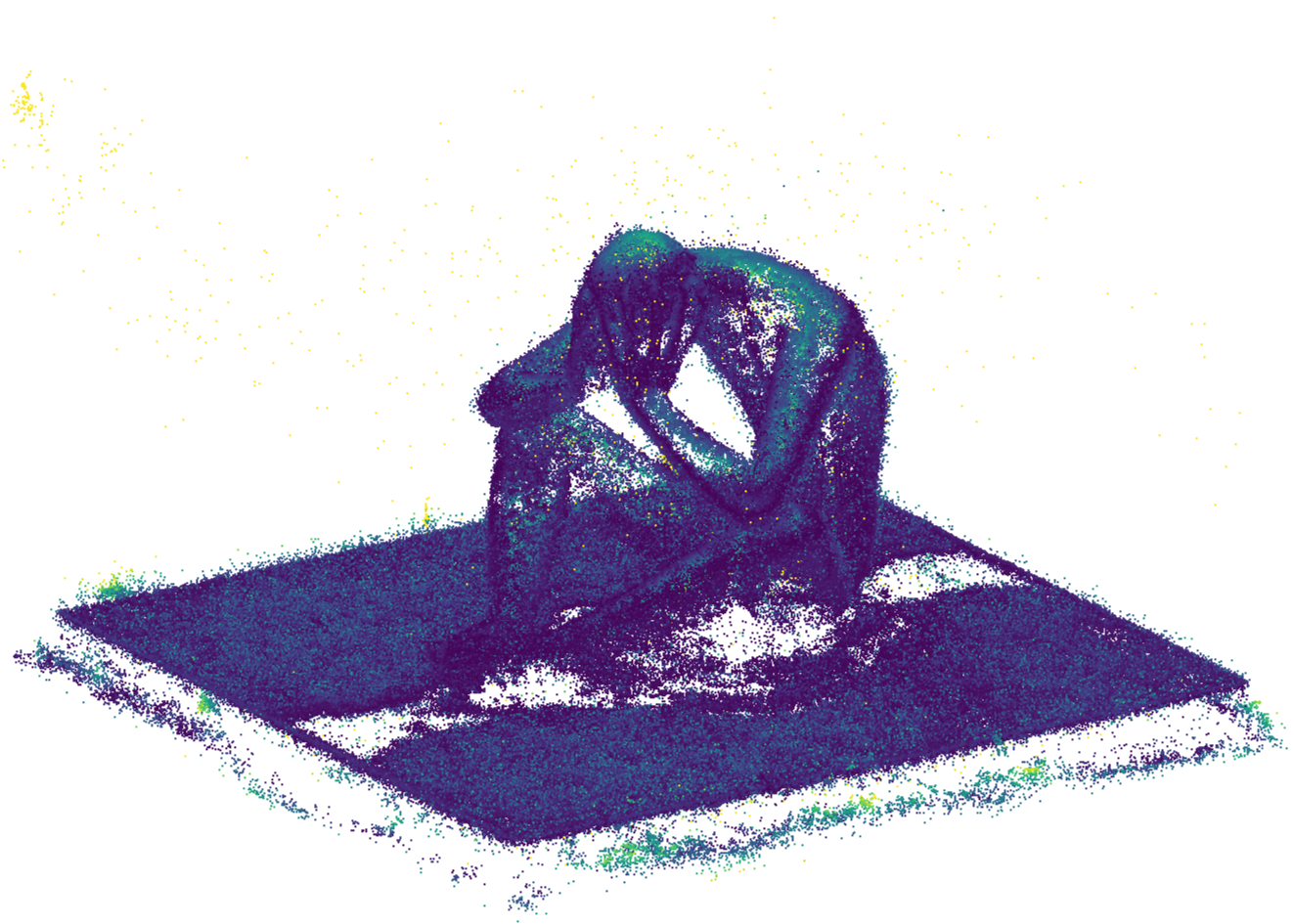}}\\
\hspace{-1.75cm}
\rotatebox{90}{internal HoloLens data}
\hspace{0.75cm}
\subfigure[]{\label{fig:Denker_Reference}
	\includegraphics[width=0.26\linewidth]{figures/Denker_Reference.png}}
	\hspace{0.2cm}
\subfigure[]{\label{fig:Denker_Reference_HoloLens}
	\includegraphics[width=0.26\linewidth]{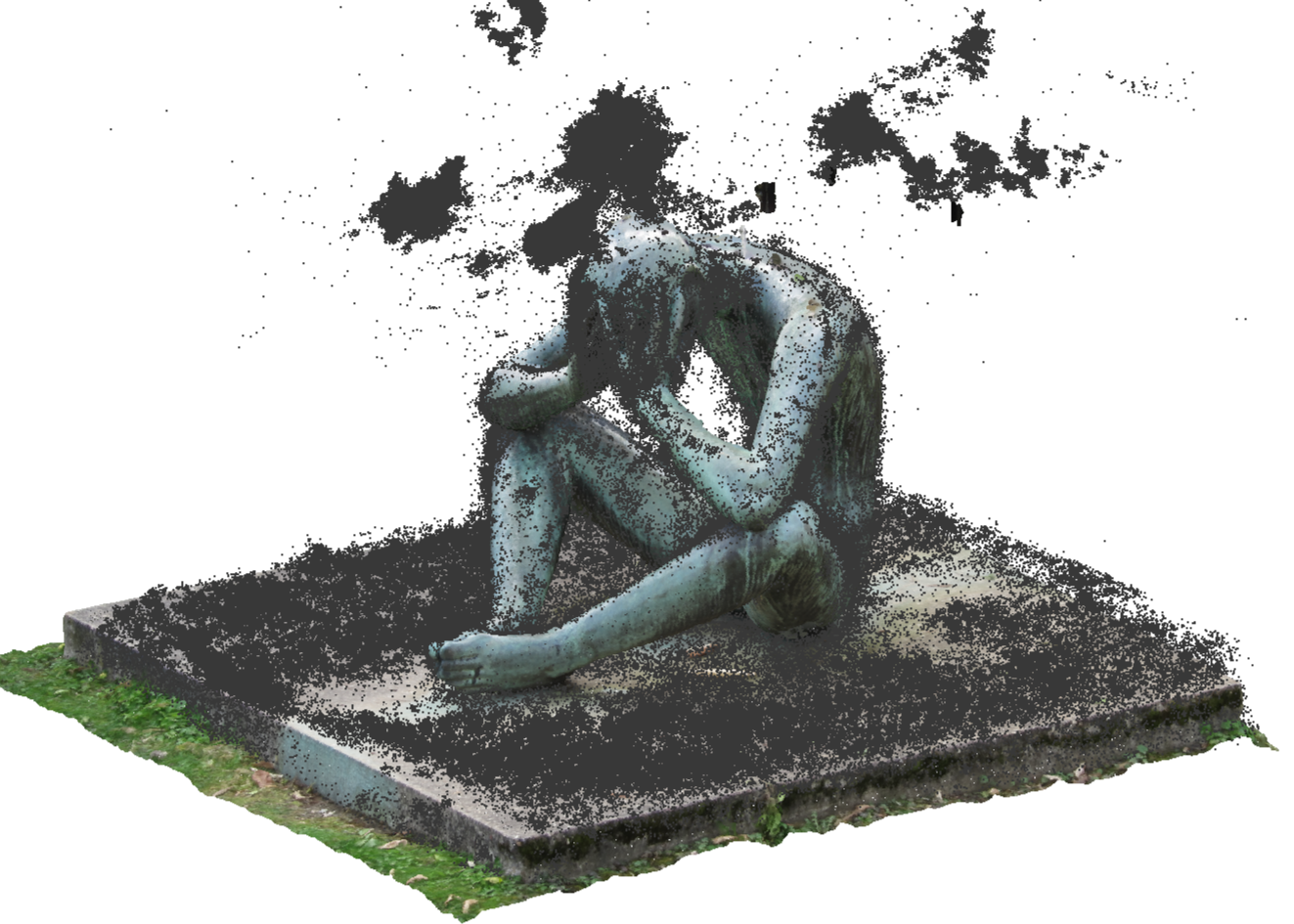}}
	\hspace{0.2cm}
\subfigure[]{\label{fig:Denker_HoloLens_C2C}  
     \includegraphics[width=0.26\linewidth]{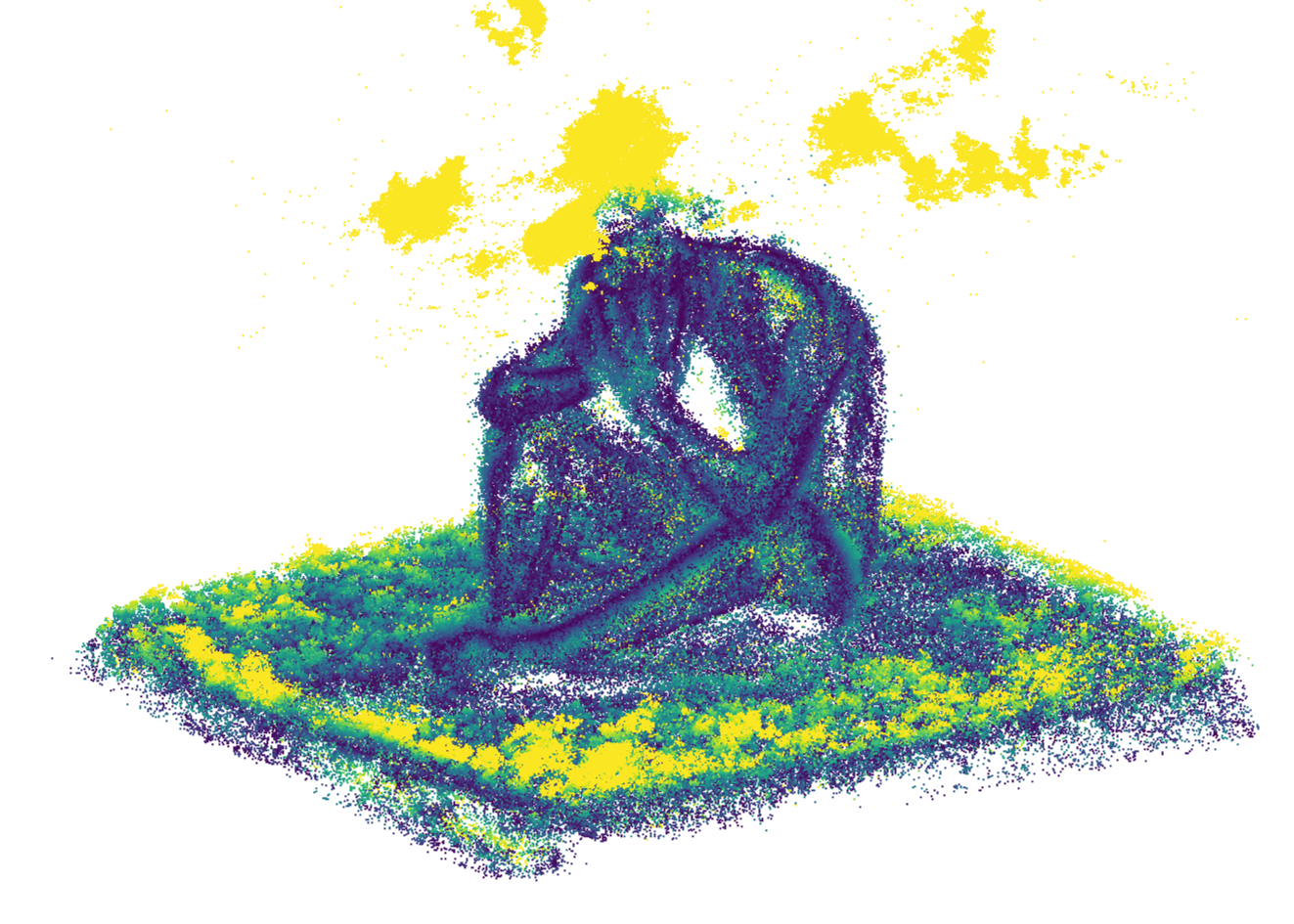}}\\
     \subfigure{\label{Denker_HoloLens_C2C}  
     \includegraphics[width=0.2\linewidth]{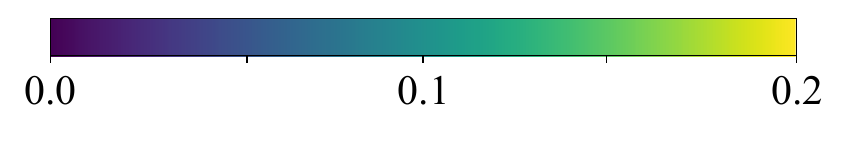}}
      \vspace{-0.25cm}
	\caption{Geometric accuracy via Chamfer Distance $\downarrow$. Reference point cloud from MVS compared to the densified point clouds from Gaussian Splatting by extracting the center of each Gaussian. Top: Scene 'Denker' with external SfM data. From left to right: \protect{\subref{fig:Denker_Reference_a}} Reference, \protect{\subref{fig:Denker_Reference_SfM}} reference and GS point cloud, \protect{\subref{fig:Denker_SfM_C2C}} Chamfer Distance of the GS point cloud. 
Bottom: Scene 'Denker' with internal HoloLens data. From left to right: \protect{\subref{fig:Denker_Reference}} Reference, \protect{\subref{fig:Denker_Reference_HoloLens}} reference and GS point cloud, \protect{\subref{fig:Denker_HoloLens_C2C}} Chamfer Distance of the GS point cloud.}
\label{fig:Denker_Chamfer}
\end{figure*}

\begin{figure*}[h!]
	\centering
\hspace{-1.75cm}
\rotatebox{90}{external SfM data}
\hspace{0.75cm}
\subfigure[]{\label{fig:Ficus_Reference_a}
	\includegraphics[width=0.26\linewidth]{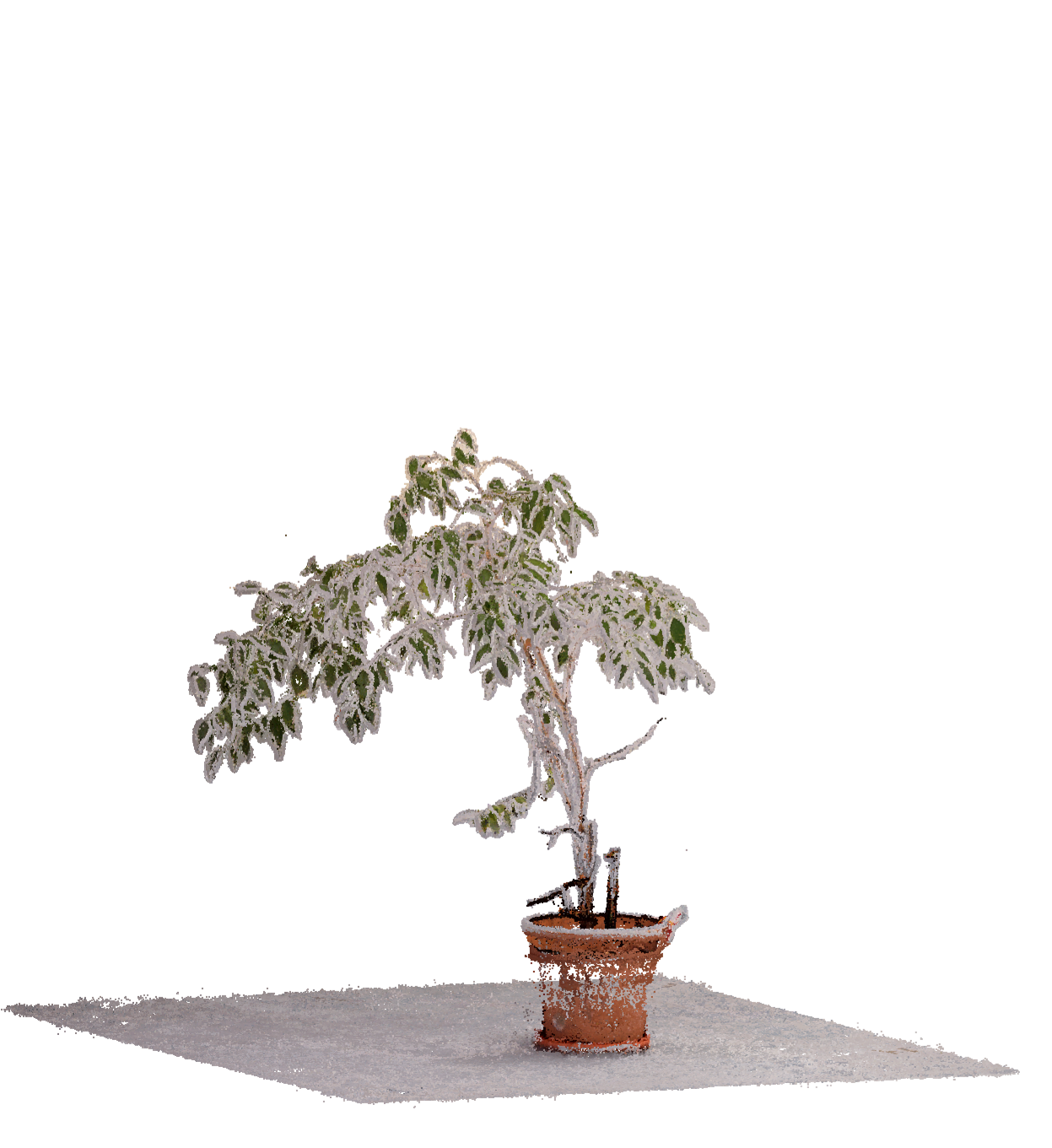}}
	\hspace{0.2cm}
\subfigure[]{\label{fig:Ficus_Reference_SfM_}
	\includegraphics[width=0.26\linewidth]{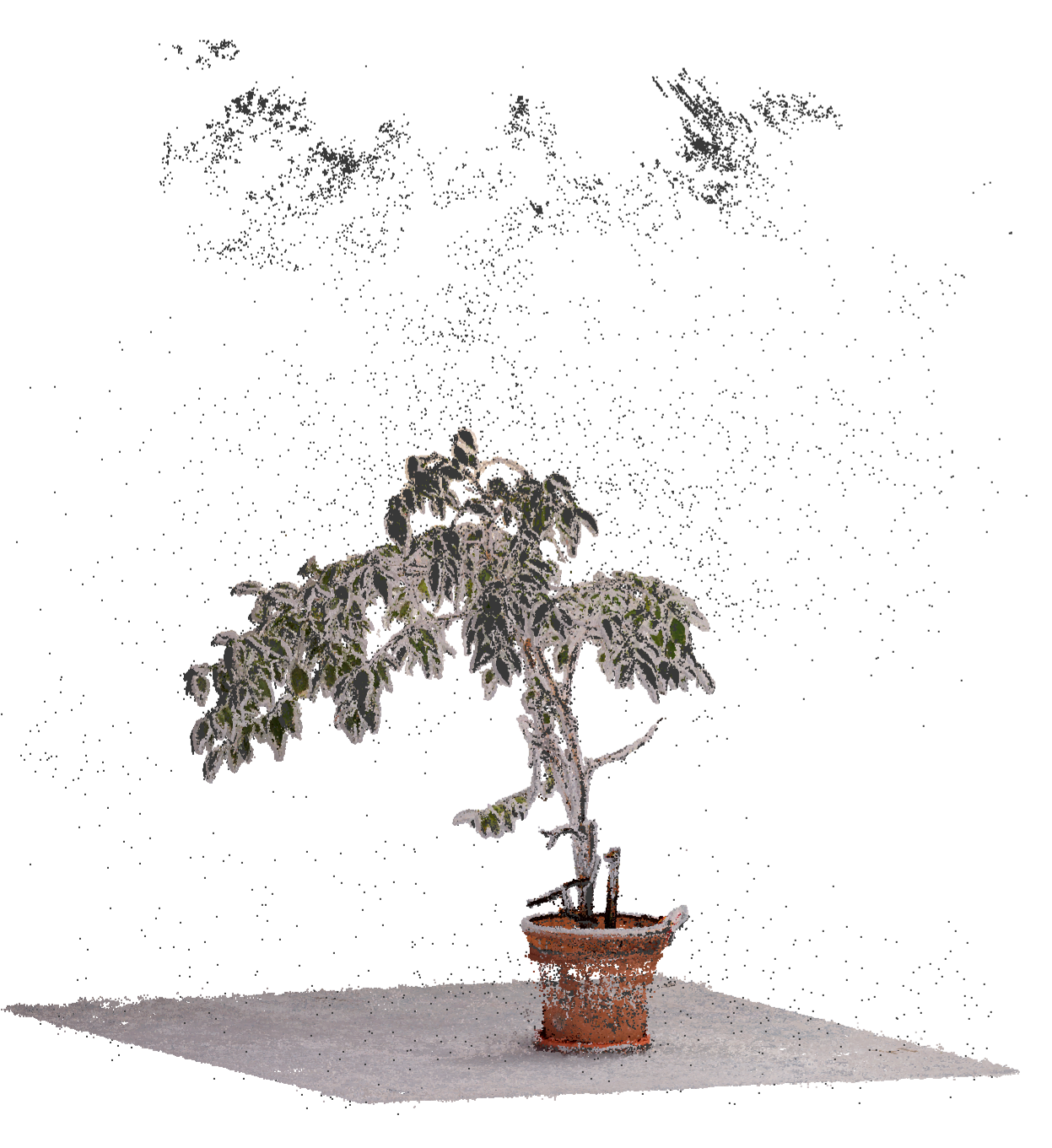}}
	\hspace{0.2cm}
\subfigure[]{\label{fig:Ficus_C2C_SfM_}  
     \includegraphics[width=0.26\linewidth]{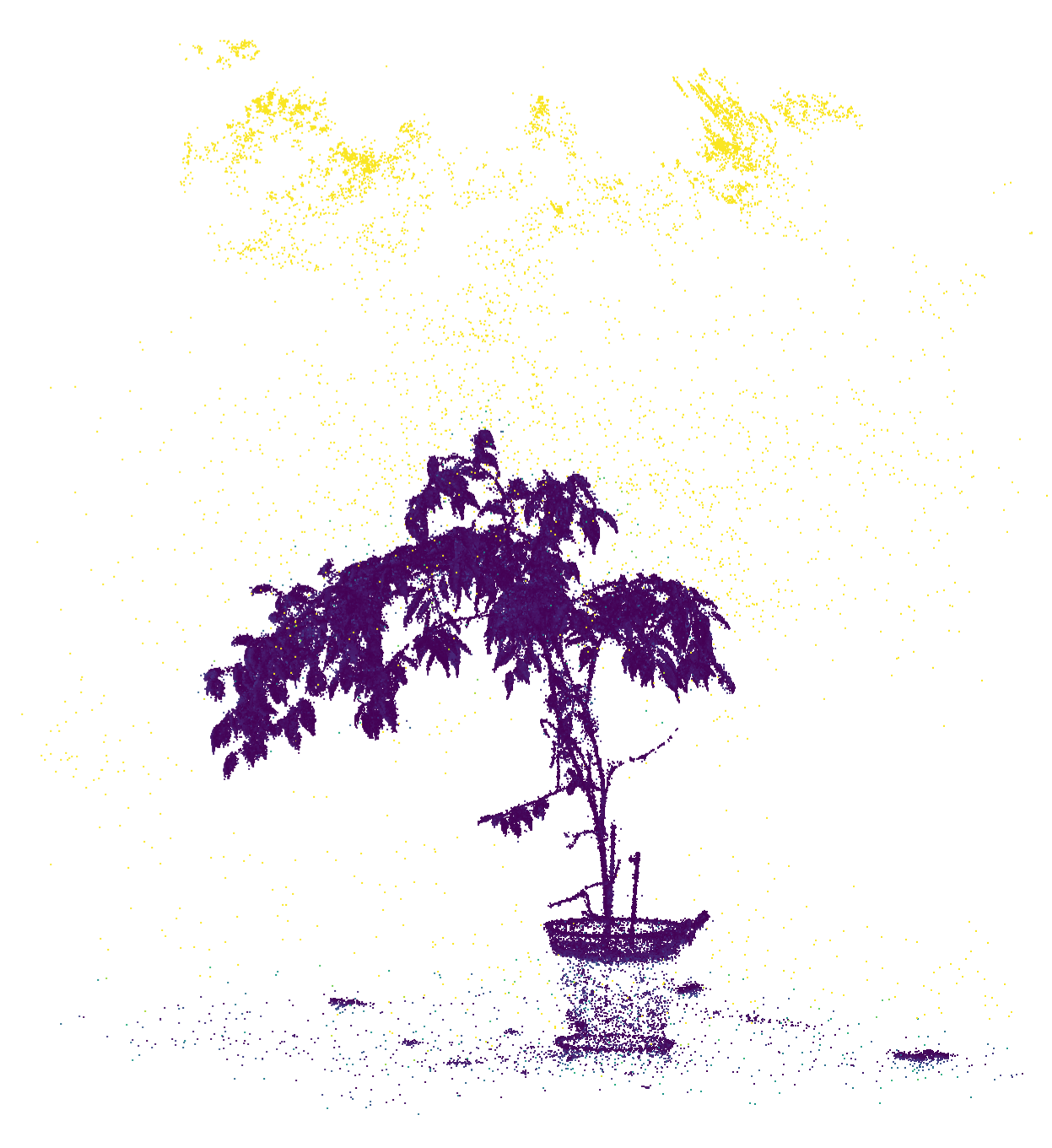}}\\
 \hspace{-1.75cm}
\rotatebox{90}{internal HoloLens data}
\hspace{0.75cm}
\subfigure[]{\label{fig:Ficus_Reference}
	\includegraphics[width=0.26\linewidth]{figures/Ficus_Reference.png}}
	\hspace{0.2cm}
\subfigure[]{\label{fig:Ficus_Reference_HoloLens_}
	\includegraphics[width=0.26\linewidth]{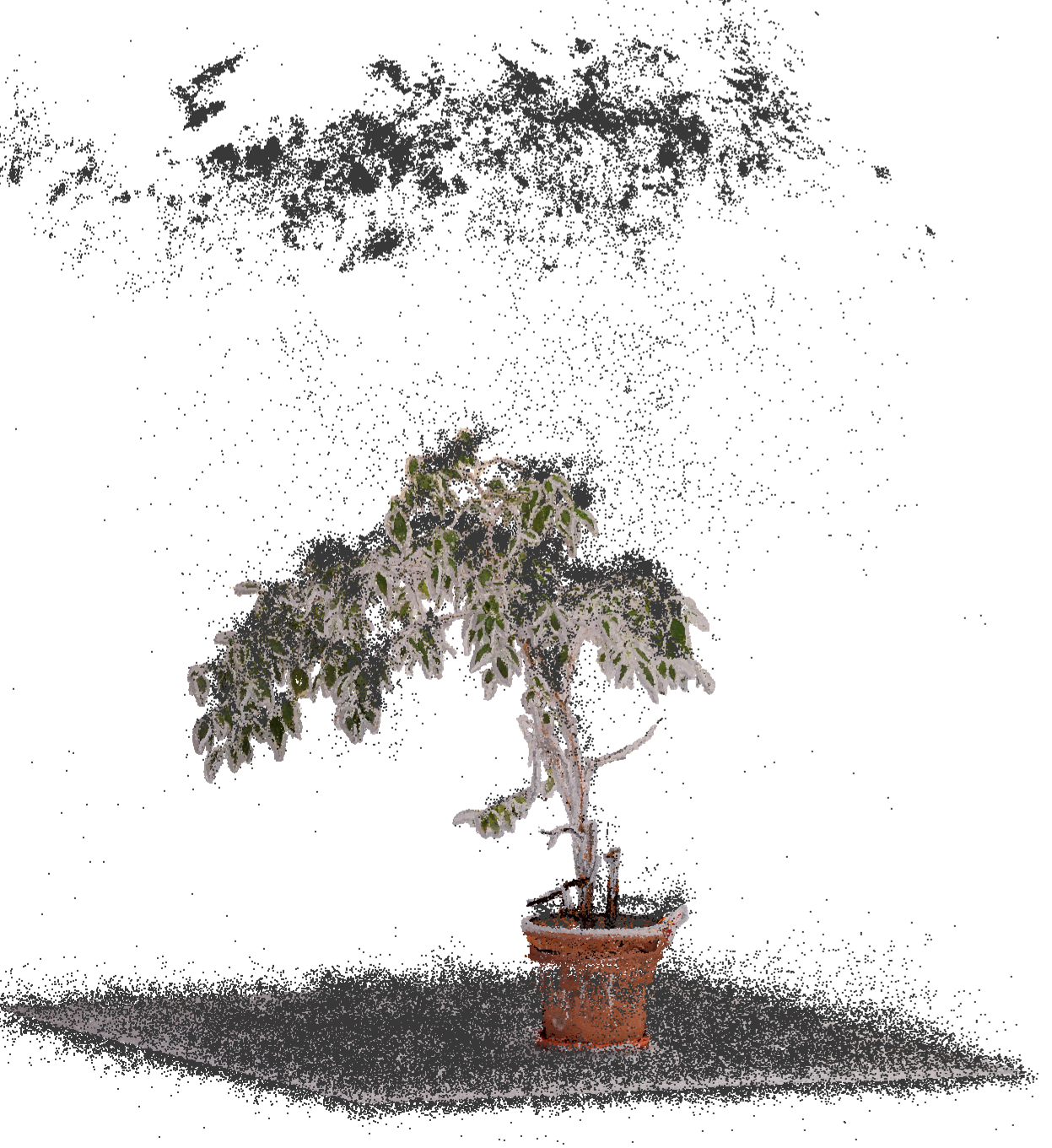}}
	\hspace{0.2cm}
\subfigure[]{\label{fig:Ficus_C2C_HoloLens_}  
     \includegraphics[width=0.26\linewidth]{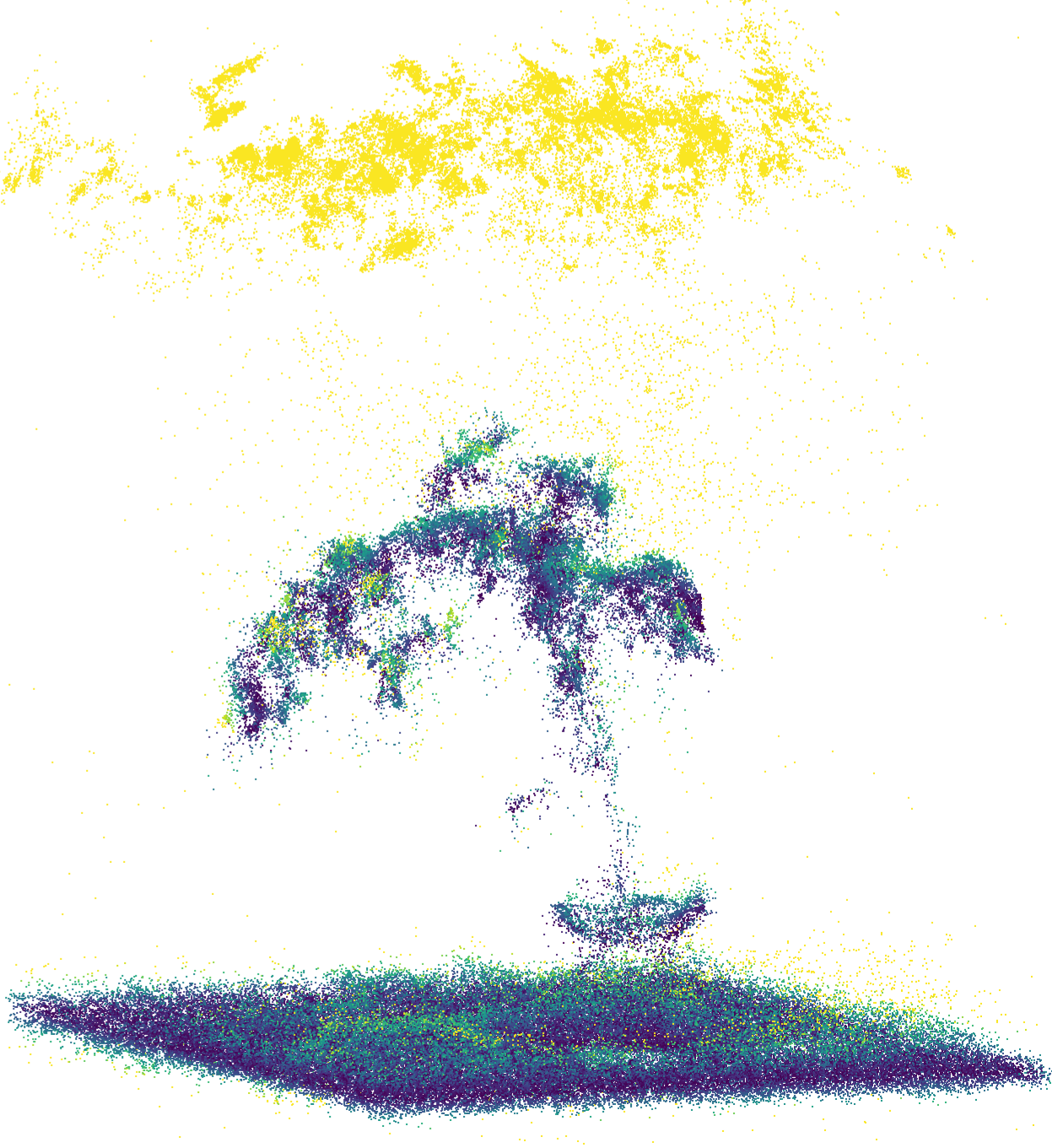}}\\
\subfigure{\label{Denker_HoloLens_C2C}  
     \includegraphics[width=0.2\linewidth]{figures/colorbar.pdf}}
      \vspace{-0.25cm}
	\caption{Geometric accuracy via Chamfer Distance $\downarrow$. Reference point cloud from MVS compared to the point clouds from Gaussian Splatting by extracting the center of each Gaussian. Top: Scene 'Ficus' with external SfM data. From left to right: \protect{\subref{fig:Ficus_Reference_a}} Reference, \protect{\subref{fig:Ficus_Reference_SfM_}} reference and GS point cloud, \protect{\subref{fig:Ficus_C2C_SfM_}} Chamfer Distance of the GS point cloud. 
Bottom: Scene 'Ficus' with internal HoloLens data. From left to right: \protect{\subref{fig:Ficus_Reference}} Reference, \protect{\subref{fig:Ficus_Reference_HoloLens_}} reference and GS point cloud, \protect{\subref{fig:Ficus_C2C_HoloLens_}} Chamfer Distance of the GS point cloud.}
\label{fig:Ficus_Chamfer}
\end{figure*}

The visual appearance of the densified point clouds is further evaluated quantitatively and qualitatively by their geometric 3D accuracy \citep{dtu} using the Chamfer cloud-to-cloud Distance from the point cloud to the reference from MVS. Consistent with the training and rendering results, the extracted point cloud exhibits similar geometric characteristics, as shown quantitatively in Table \ref{tab:Chamfer} and qualitatively in Figures \ref{fig:Denker_Chamfer} and \ref{fig:Ficus_Chamfer}. \newline
The quantitative results in Table \ref{tab:Chamfer} highlight clear differences in the geometric accuracy of the extracted densified point clouds, especially regarding the two types of initial input data.
For the scene 'Denker', with a mean accuracy of 0.021 and a standard deviation of 0.061 for external SfM data, contrasting with a significantly lower accuracy of 0.298 and a standard deviation of 0.534 for internal HoloLens data.
Similar results are obtained for the scene 'Ficus', where the use of external SfM data shows a mean geometric accuracy of 0.045, with a standard deviation of 0.261. In contrast, the use of internal HoloLens data again results in a significantly lower accuracy of 0.596, with a standard deviation of 0.891.
The point clouds, illustrating the Chamfer Distances, visually underscore the quantitative findings, as shown in Figure \ref{fig:Denker_Chamfer} for the scene 'Denker'. It is evident that the external SfM data yields a high accuracy for the statue's surface, although a lower geometric accuracy is noticeable on low-textured, homogeneous areas. The same trend is seen for the internal HoloLens data, where lower geometric accuracy is reflected in noisy edges and floater artifacts. Figure \ref{fig:Ficus_Chamfer} shows a similar pattern for the scene 'Ficus'. The SfM data performs well in capturing details, while the HoloLens data shows reduced accuracy, especially on fine-structured object parts like the branches.

\begin{table}[h!]
\centering
\begin{tabular}{|l|c|c|c|c|}\hline
 &\multicolumn{2}{c|}{External SfM data} & \multicolumn{2}{c|}{Internal HoloLens data}\\\hline
& mean & std & mean & std\\ \hline
Denker &0.021&0.061 & 0.298 & 0.534\\
Ficus&0.045&0.261 & 0.596 & 0.891\\\hline
\end{tabular}
   \vspace{-0.2cm}
\caption{Geometric 3D accuracy via Chamfer Distance $\downarrow$. Mean distance (mean) and standard derivation (std) for the scenes 'Denker' and 'Ficus' each with external SfM and internal HoloLens data. Note that the reference point clouds and therefore the Chamfer Distances are non-metrical.}
\label{tab:Chamfer}
\end{table}

\section{Discussion}
In this paper, we introduce HoloGS to investigate the application of instant 3D Gaussian Splatting using data from the internal sensors of the Microsoft HoloLens 2. Specifically, the input data consists of RGB images with corresponding camera poses and a point cloud from the depth data of the HoloLens as initial Gaussian centers. 
We have shown both quantitatively and qualitatively that the internal HoloLens data is suitable for this application as it enables convergence of the Gaussian Splatting optimization process. 
The optimization of the training process based on the rendered RGB images converges quickly and reaches a maximum PSNR value of 20.17 for the scene 'Denker' and 20.55 for 'Ficus'. This convergence enables the rendering of novel synthetic images from different views which represent the scene visually well. Additionally, the optimization of the Gaussians during the training process enables the extraction of the densified point cloud from the Gaussian centers.

Nevertheless, limitations exist, as the results of the externally preprocessed SfM data outperform the results of the internal HoloLens data. Thereby, a higher maximum PSNR is reached during the training, and the rendered images appear less blurry and contain less floater artifacts in comparison to the internal HoloLens data. In addition, the geometric accuracy of the 3D reconstruction of the densified point clouds with HoloLens performs 10 times weaker on average, although this may be due to floater artifacts, which weigh heavily.
We suspect the cause of these discrepancies lies in the less precise camera poses of the RGB images of the internal HoloLens data, leading to blurry results and artifacts in the rendered images as well as the densified point clouds. Moreover, it could be assumed that the initial point cloud from the depth sensor may not match the correct positions of the RGB images in the coordinate system. However, since the PSNR does not continue to increase during training as Gaussians grow, shrink or are removed, we reject this assumption as a potential cause. 

Therefore, firstly, considering the 3D mapping aspect, the usage of HoloLens solely as a 3D mapping system, without the 3D scene reconstruction aspect of computer graphics, a direct usage of the depth maps and the resulting point cloud from the HoloLens sensor data can be considered. This results in a point cloud with high point density and fine details, as shown by the input data (Section \ref{sec:dataset}) of the internal HoloLens data approach.
Secondly, from the aspect of computer graphics, we nonetheless consider the combination of HoloLens and Gaussian Splatting to be suitable.  
If, as suspected, the weaker results are due to the RGB camera poses, just as with the HoloLens-NeRF combination \citep{hololens_nerf_jaeger}, we propose the optimization of RGB camera poses during the training process. A strategy previously employed in the context of NeRF and generally beneficial for low-quality or unknown camera poses \citep{Barf, CBARF, Garf, instantngp_refinement, GNerf, NoPe-NeRF}.
By optimizing the camera poses during the training, the quality of the rendering and the densified point cloud from Gaussian Splatting could reach the quality of the SfM without additional pre-processing time. Furthermore, the real-time capability of the HoloLens offers the potential to insert data into Gaussian Splatting during the optimization, which is, with regard to SLAM approaches \citep{slam1, slam2, slam3}, quite appealing.


\begin{figure}[H]
 \vspace{-0.2cm}
	\centering
\subfigure[]{\label{fig:untex_input_sfm}
	\includegraphics[width=0.37\columnwidth]{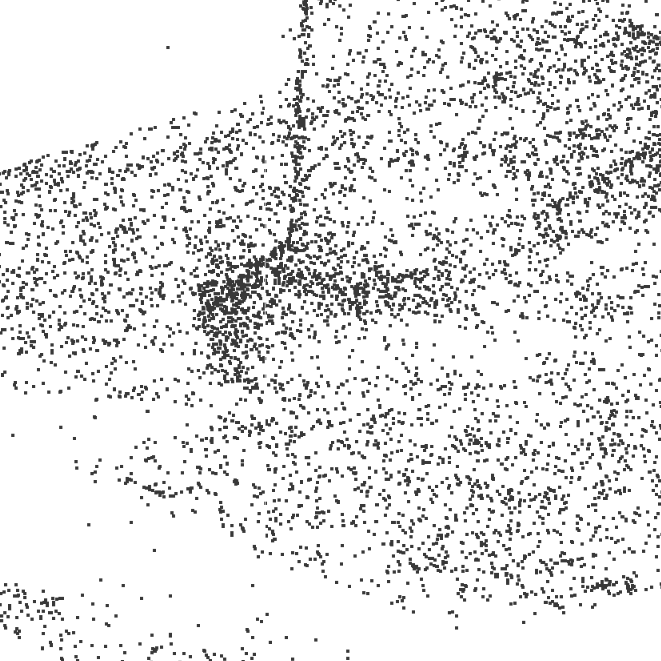}}
  \hspace{0.2cm}
\subfigure[]{\label{fig:untex_output_sfm}
	\includegraphics[width=0.37\columnwidth]{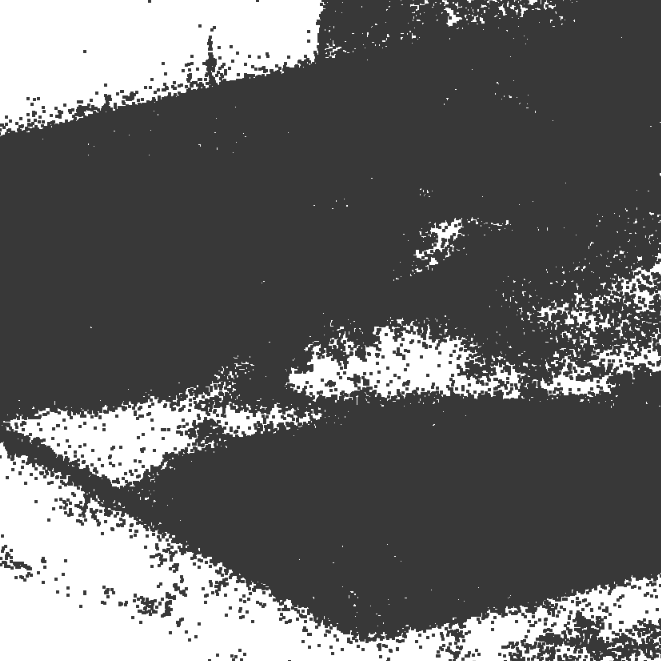}}\\
  \vspace{-0.35cm}
\subfigure[]{\label{fig:untex_rendered_sfm}
	\includegraphics[width=0.37\columnwidth]{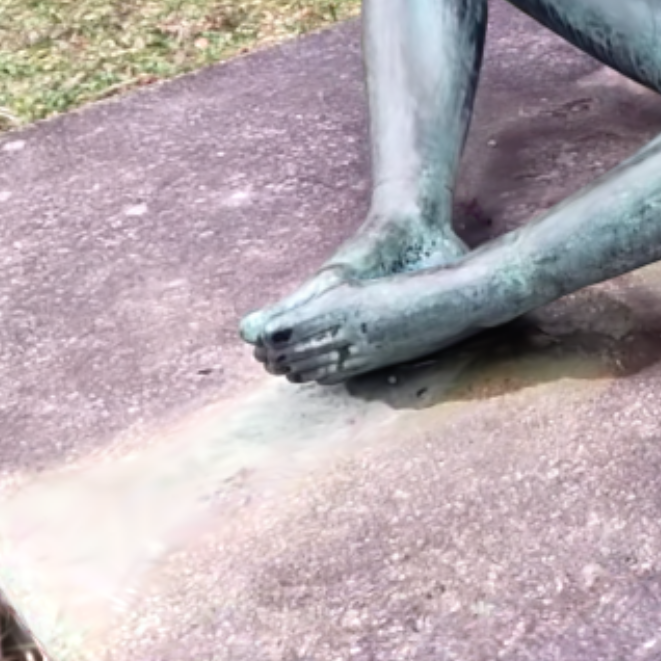}}
	  \hspace{0.2cm}
\subfigure[]{\label{fig:untex_ref}
	\includegraphics[width=0.37\columnwidth]{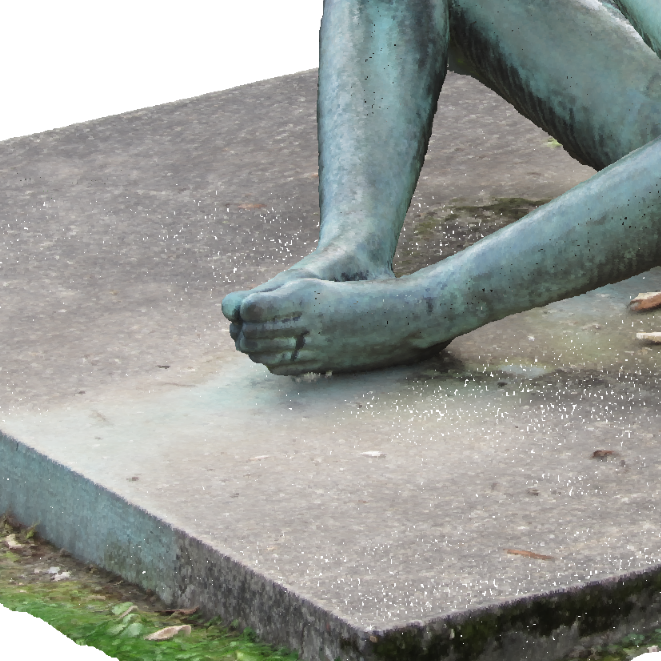}}
 \vspace{-0.5cm}
	\caption{Low-textured, homogeneous surfaces. \protect{\subref{fig:untex_input_sfm}} Input SfM point cloud whose points are used as initial Gaussian centers. \protect{\subref{fig:untex_output_sfm}}
Output densified point cloud of the Gaussian centers after training. \protect{\subref{fig:untex_rendered_sfm}} Rendered image. \protect{\subref{fig:untex_ref}} Reference point cloud. It can be observed that the homogeneous, low-textured surfaces clearly have a lower point density of Gaussians.}
\label{fig:Untextured}
\end{figure}
Generally, gaps in the densified point cloud, for both SfM and HoloLens data, are present, as shown in Figure \ref{fig:Untextured}. These gaps probably result from areas with homogeneous color, such as the areas around the legs of the statue in the scene 'Denker' and the low-textured pot of the plant in the scene 'Ficus'. Therefore, for the densified point cloud extraction, simply extracting the Gaussian centers is insufficient, due to the presence of floater artifacts and non-uniform point density on low-textured surfaces, where only a few individual Gaussians exist for homogeneous colors. In addition, it is not yet clear whether the Gaussian center or the Gaussian surface best represents the surface of the object geometry. These issues can be resolved by a suitable method for 3D point cloud extraction beyond querying of the Gaussian centers by further post-processing steps and extensions. In summary, despite the challenges, we see potential for HoloGS by combining the Microsoft HoloLens 2 with Gaussian Splatting for an instant 3D scene reconstruction and point cloud extraction.

\section{Conclusion}
In conclusion, HoloGS enables instant 3D scene reconstruction with 3D Gaussian Splatting directly from the internal sensor data of the Microsoft HoloLens 2. Although our results show some weaknesses compared to more elaborate approaches of using externally computed SfM data, such as a lower PSNR, floater artifacts and blurring in the rendered images as well as artifacts in the extracted point cloud, we nevertheless see potential for further optimization. In particular, refining the RGB camera poses during the training process could improve the results and enable real-time 3D reconstruction using state-of-the-art methods and entertainment devices like the HoloLens. As well as additional methods for point cloud and surface reconstruction from Gaussian Splatting.
HoloGS thus represents a promising solution for using the Microsoft HoloLens 2 for instant 3D Gaussian Splatting, which offers further research potential in the realm of photogrammetry, computer vision and computer graphics.

\section*{ACKNOWLEGDEMENTS}\label{ACKNOWLEGDEMENTS}
The authors would like to thank Patrick Hübner from Technical University of Darmstadt for support regarding the HoloLens.

\bibliographystyle{elsarticle-harv}
{
	\begin{spacing}{1.}
		\normalsize
		\bibliography{ISPRSguidelines_authors} 

\begin{thebibliography}{xx}

\bibitem[Bian et al., 2023]{NoPe-NeRF}
Bian, W., Wang, Z., Li, K., Bian, J.-W., Prisacariu, V.~A., 2023.
 Nope-nerf: Optimising neural radiance field with no pose prior.
 \emph{Proceedings of the IEEE/CVF Conference on Computer Vision and Pattern
  Recognition (CVPR)}, 4160--4169.

\bibitem[Chng et al., 2022]{Garf}
Chng, S.-F., Ramasinghe, S., Sherrah, J., Lucey, S., 2022.
 Gaussian activated neural radiance fields for high fidelity reconstruction and
  pose estimation.
 \emph{Computer Vision--ECCV 2022: 17th European Conference, Tel Aviv, Israel,
  October 23--27, 2022, Proceedings, Part XXXIII}, Springer, 264--280.

\bibitem[Darmon et al., 2022]{NeuralWarp}
Darmon, F., Bascle, B., Devaux, J.-C., Monasse, P., Aubry, M., 2022.
 Improving neural implicit surfaces geometry with patch warping.
 \emph{Proceedings of the IEEE/CVF Conference on Computer Vision and Pattern
  Recognition}, 6260--6269.

\bibitem[Dibene and Dunn, 2022]{HoloLens2_Streaming}
Dibene, J.~C., Dunn, E., 2022.
 HoloLens 2 Sensor Streaming.
 {\em arXiv}.
 https://arxiv.org/abs/2211.02648.

\bibitem[Fink et al., 2023]{slam1}
Fink, L., R{\"u}ckert, D., Franke, L., Keinert, J., Stamminger, M., 2023.
 Livenvs: Neural view synthesis on live rgb-d streams.
 \emph{SIGGRAPH Asia 2023 Conference Papers}, 1--11.

\bibitem[Fu et al., 2023]{CBARF}
Fu, H., Yu, X., Li, L., Zhang, L., 2023.
 Cbarf: Cascaded bundle-adjusting neural radiance fields from imperfect camera
  poses.
 arXiv: 2310.09776.

\bibitem[Haitz et al., 2023]{hololens_dennis}
Haitz, D., Jutzi, B., Ulrich, M., J\"ager, M., H\"ubner, P., 2023.
 Combining HoloLens with Instant-NeRFs: Advanced real-time 3D Mobile Mapping.
 {\em The International Archives of the Photogrammetry, Remote Sensing and
  Spatial Information Sciences}, XLVIII-1/W1-2023, 167--174.
 https://isprs-archives.copernicus.org/articles/XLVIII-1-W1-2023/167/2023/.

\bibitem[Hou et al., 2024]{hololens_iphone}
Hou, J., Hübner, P., Schmidt, J., Iwaszczuk, D., 2024.
 Indoor Mapping with Entertainment Devices: Evaluating the Impact of Different
  Mapping Strategies for Microsoft HoloLens 2 and Apple iPhone 14 Pro.
 {\em Sensors}, 24(4).

\bibitem[J\"ager et al., 2023]{hololens_nerf_jaeger}
J\"ager, M., H\"ubner, P., Haitz, D., Jutzi, B., 2023.
 A Comparative Neural Radiance Field (NeRF) 3D Analysis of Camera Poses from
  HoloLens Trajectories and Structure from Motion.
 {\em The International Archives of the Photogrammetry, Remote Sensing and
  Spatial Information Sciences}, XLVIII-1/W1-2023, 207--213.
 https://isprs-archives.copernicus.org/articles/XLVIII-1-W1-2023/207/2023/.

\bibitem[J\"ager and Jutzi, 2023]{densitygradient_jaeger}
J\"ager, M., Jutzi, B., 2023.
 3D Density-Gradient based Edge Detection on Neural Radiance Fields (NeRFs) for
  Geometric Reconstruction.
 {\em The International Archives of the Photogrammetry, Remote Sensing and
  Spatial Information Sciences}, XLVIII-1/W3-2023, 71--78.
 https://isprs-archives.copernicus.org/articles/XLVIII-1-W3-2023/71/2023/.

\bibitem[Jensen et al., 2014]{dtu}
Jensen, R., Dahl, A., Vogiatzis, G., Tola, E., Aan{\ae}s, H., 2014.
 Large scale multi-view stereopsis evaluation.
 \emph{2014 IEEE Conference on Computer Vision and Pattern Recognition}, IEEE,
  406--413.

\bibitem[Jäger et al., 2023]{Jaeger_Nerfensemble}
Jäger, M., Landgraf, S., Jutzi, B., 2023.
 Density uncertainty quantification with nerf-ensembles: Impact of data and
  scene constraints.

\bibitem[Kerbl et al., 2023]{kerbl3Dgaussians}
Kerbl, B., Kopanas, G., Leimk{\"u}hler, T., Drettakis, G., 2023.
 3D Gaussian Splatting for Real-Time Radiance Field Rendering.
 {\em ACM Transactions on Graphics}, 42(4).
 https://repo-sam.inria.fr/fungraph/3d-gaussian-splatting/.

\bibitem[Li et al., 2023]{neuralangelo}
Li, Z., M\"uller, T., Evans, A., Taylor, R.~H., Unberath, M., Liu, M.-Y., Lin,
  C.-H., 2023.
 Neuralangelo: High-fidelity neural surface reconstruction.
 \emph{IEEE Conference on Computer Vision and Pattern Recognition ({CVPR})}.

\bibitem[Lin et al., 2021]{Barf}
Lin, C.-H., Ma, W.-C., Torralba, A., Lucey, S., 2021.
 Barf: Bundle-adjusting neural radiance fields.
 \emph{Proceedings of the IEEE/CVF International Conference on Computer
  Vision}, 5741--5751.

\bibitem[Lin et al., 2023]{instantngp_refinement}
Lin, Y., Müller, T., Tremblay, J., Wen, B., Tyree, S., Evans, A., Vela, P.~A.,
  Birchfield, S., 2023.
 Parallel inversion of neural radiance fields for robust pose estimation.
 \emph{2023 IEEE International Conference on Robotics and Automation (ICRA)},
  9377--9384.

\bibitem[Lowe, 2004]{Lowe}
Lowe, D.~G., 2004.
 Distinctive image features from scale-invariant keypoints.
 {\em International journal of computer vision}, 60, 91--110.

\bibitem[Meng et al., 2021]{GNerf}
Meng, Q., Chen, A., Luo, H., Wu, M., Su, H., Xu, L., He, X., Yu, J., 2021.
 Gnerf: Gan-based neural radiance field without posed camera.
 \emph{Proceedings of the IEEE/CVF International Conference on Computer
  Vision}, 6351--6361.

\bibitem[Mildenhall et al., 2020]{mildenhall_et_al_2020}
Mildenhall, B., Srinivasan, P.~P., Tancik, M., Barron, J.~T., Ramamoorthi, R.,
  Ng, R., 2020.
 {NeRF: Representing Scenes as Neural Radiance Fields for View Synthesis}.
 \emph{European Conference on Computer Vision (ECCV)}, 405--421.

\bibitem[Oechsle et al., 2021]{unisurf}
Oechsle, M., Peng, S., Geiger, A., 2021.
 Unisurf: Unifying neural implicit surfaces and radiance fields for multi-view
  reconstruction.
 \emph{Proceedings of the IEEE/CVF International Conference on Computer
  Vision}, 5589--5599.

\bibitem[Park et al., 2019]{deepsdf}
Park, J.~J., Florence, P., Straub, J., Newcombe, R., Lovegrove, S., 2019.
 Deepsdf: Learning continuous signed distance functions for shape
  representation.
 \emph{Proceedings of the IEEE/CVF conference on computer vision and pattern
  recognition}, 165--174.

\bibitem[Rosinol et al., 2023]{slam2}
Rosinol, A., Leonard, J.~J., Carlone, L., 2023.
 Nerf-slam: Real-time dense monocular slam with neural radiance fields.
 \emph{2023 IEEE/RSJ International Conference on Intelligent Robots and Systems
  (IROS)}, IEEE, 3437--3444.

\bibitem[Sch{\"{o}}nberger and Frahm, 2016]{Schonberger_2016_CVPR}
Sch{\"{o}}nberger, J.~L., Frahm, J.-M., 2016.
 Structure-from-motion revisited.
 \emph{Proceedings of the IEEE Conference on Computer Vision and Pattern
  Recognition (CVPR)}.

\bibitem[Wang et al., 2021]{NeuS}
Wang, P., Liu, L., Liu, Y., Theobalt, C., Komura, T., Wang, W., 2021.
 Neus: Learning neural implicit surfaces by volume rendering for multi-view
  reconstruction.
 ~34, 27171--27183.

\bibitem[Weinmann et al., 2020]{jaeger}
Weinmann, M., J{\"a}ger, M.~A., Wursthorn, S., Jutzi, B., H{\"u}bner, P., 2020.
 3D indoor mapping with the Microsoft HoloLens: qualitative and quantitative
  evaluation by means of geometric features.
 {\em ISPRS Annals of the Photogrammetry, Remote Sensing and Spatial
  Information Sciences}, 1, 165--172.

\bibitem[Weinmann et al., 2021]{hololens_weinmann}
Weinmann, M., Wursthorn, S., Weinmann, M., H{\"{u}}bner, P., 2021.
 Efficient 3D Mapping and Modelling of Indoor Scenes with the Microsoft
  HoloLens: A Survey.
 {\em Journal of photogrammetry, remote sensing and geoinformation science},
  89, 319-333.

\bibitem[Yariv et al., 2021]{volsdf}
Yariv, L., Gu, J., Kasten, Y., Lipman, Y., 2021.
 Volume rendering of neural implicit surfaces.
 {\em Advances in Neural Information Processing Systems}, 34, 4805--4815.

\bibitem[Zhang et al., 2022]{regSDF}
Zhang, J., Yao, Y., Li, S., Fang, T., McKinnon, D., Tsin, Y., Quan, L., 2022.
 Critical regularizations for neural surface reconstruction in the wild.
 \emph{Proceedings of the IEEE/CVF Conference on Computer Vision and Pattern
  Recognition}, 6270--6279.

\bibitem[Zhu et al., 2022]{slam3}
Zhu, Z., Peng, S., Larsson, V., Xu, W., Bao, H., Cui, Z., Oswald, M.~R.,
  Pollefeys, M., 2022.
 Nice-slam: Neural implicit scalable encoding for slam.
 \emph{Proceedings of the IEEE/CVF Conference on Computer Vision and Pattern
  Recognition}, 12786--12796.

\end{thebibliography}
	\end{spacing}
}

\end{document}